\documentclass{tADR2e}

\usepackage{graphicx}
\usepackage{subfigure}
\usepackage{algorithm}
\usepackage{algorithmic}
\usepackage{amsthm}
\newtheorem{prop}{Proposition}

\theoremstyle{plain}

\theoremstyle{definition}

\theoremstyle{remark}

\begin{document}

\jvol{00} \jnum{00} \jyear{2013} \jmonth{January}

\articletype{FULL PAPER}

\title{Situated GAIL: Multitask imitation using task-conditioned\\adversarial inverse reinforcement learning}

\author{Kyoichiro Kobayashi$^{a}$,
Takato Horii$^{b,c}$$^{\ast}$\thanks{$^\ast$Corresponding author. Email: takato@sys.es.osaka-u.ac.jp \vspace{6pt}}, 
Ryo Iwaki$^{a}$, Yukie Nagai$^{c}$ and Minoru Asada$^{d}$ \\ \vspace{6pt}
$^{a}${\em{Graduate School of Engineering, Osaka University, Osaka, Japan}}; \\
$^{b}${\em{Graduate School of Engineering Science, Osaka University, Osaka, Japan}}; \\
$^{c}${\em{International Research Center for Neurointelligence, The University of Tokyo, Tokyo, Japan}}; \\
$^{d}${\em{Institute for Open and Transdisciplinary Research Initiatives, Osaka University, Osaka, Japan}} \\\vspace{6pt}
\received{v1.0 released January 2013} }

\maketitle

\begin{abstract}
Generative adversarial imitation learning (GAIL) has attracted increasing attention in the field of robot learning.
It enables robots to learn a policy to achieve a task demonstrated by an expert while simultaneously estimating the reward function behind the expert's behaviors.
However, this framework is limited to learning a single task with a single reward function.  
This study proposes an extended framework called situated GAIL (S-GAIL), in which a task variable is introduced to both the discriminator and generator of the GAIL framework.
The task variable has the roles of discriminating different contexts and making the framework learn different reward functions and policies for multiple tasks.
To achieve the early convergence of learning and robustness during reward estimation, we introduce a term to adjust the entropy regularization coefficient in the generator's objective function.
Our experiments using two setups (navigation in a discrete grid world and arm reaching in a continuous space) demonstrate that the proposed framework can acquire multiple reward functions and policies more effectively than existing frameworks.
The task variable enables our framework to differentiate contexts while sharing common knowledge among multiple tasks.

\medskip

\end{abstract}

\section{Introduction}
Intelligent agents such as robots need decision-making rules to generate desired behaviors.
However, learning such rules through self-exploration requires enormous amounts of time and effort.
A promising way to facilitate learning is to learn from experts.
If experts' behavior data can be obtained through observation, robots can learn a behavioral strategy more effectively.
Hence, imitation learning and learning from demonstration have been proposed as such techniques \cite{Argall2009,Attia2018,Mueller2018robust, Lynch2019learning}, and their validity has been successfully demonstrated in various applications, for instance, navigation \cite{Pomerleau1991, Ziebart2008}, autonomous driving \cite{Ross2009, Abbeel_Heri2007}, object manipulation \cite{Finn_CB2016, Xie2019improvisation}, and so on.
\par
Imitation learning involves two issues: the first is to estimate what an expert tries to achieve (i.e., the goal), and the second is to learn how to achieve the estimated goal (i.e., the means).
We assume that there are several ways to achieve the goal and that a robot learner can only observe a limited number of demonstrations by an expert.
Thus, it is insufficient for the robot to just copy and interpolate/extrapolate the observed expert's behaviors.
Instead, the robot is expected to infer a decision-making rule from the expert to generate appropriate behaviors even in unknown situations. 
Researchers have formulated the above two issues using the framework of reinforcement learning \cite{Argall2009}.
The first issue (i.e., inferring the goal) is considered as inverse reinforcement learning (IRL) \cite{Russell1998}, by which a robot estimates a reward function behind an expert's behaviors.
The second issue (i.e., learning the means) corresponds to reinforcement learning (RL) \cite{Sutton2018}, by which the robot learns a policy to maximize the future reward from the estimated reward function.
The IRL and RL algorithms together enable a robot to achieve higher robustness than copying expert behaviors by generalizing the policy to new states and actions.
\par
Ho and Ermon \cite{Ho_GAIL2016} proposed a framework called generative adversarial imitation learning (GAIL) based on the above formulation. 
Compared to other frameworks that deal with the IRL and RL problems sequentially, GAIL aims to solve these two problems simultaneously.
The GAIL framework consists of a generator and discriminator and makes them learn in an adversarial manner, as for generative adversarial network (GAN) \cite{Goodfellow2014} does.
The generator learns to produce desired behaviors, while the discriminator learns to discriminate the output of the generator from the expert's behaviors.
In this way, the generator acquires a policy to produce optimal behaviors, which cannot be differentiated from the expert's behaviors.
\par
Following the success of GAIL, several extensions of GAIL have been proposed to take advantage of its learning efficiency and model-free characteristics.
For example, InfoGAIL \cite{Li_infoGAIL2017} introduced a latent variable to GAIL to represent multiple policies of an expert. 
This variable is used as an additional input to the generator and is optimized by maximizing the mutual information between the latent variable and the generator's output.
After learning, the variable works as an intention to switch multiple policies. 
\par
However, there are common limitations in GAIL, InfoGAIL, and related models.
First, the discriminator of GAIL and InfoGAIL is not suitable for estimating a reward function.
Instead, they learn policies without explicitly representing a reward.
This issue was noted in \cite{Fu_AIRL2017}.
The authors suggested that recovering a reward function can achieve higher robustness because learned policies can be disentangled from the environment dynamics. 
To address the issue, the authors \cite{Fu_AIRL2017} proposed a new framework called adversarial IRL (AIRL) with a modified structure of the discriminator.
Second, existing frameworks assume that an expert has a single goal and thus a single reward function governing the expert's behaviors.
Therefore, a learning module must be duplicated if the expert demonstrates multiple tasks with different goals.
An open challenge is to design a new framework that can learn multiple tasks within a single learning module.
If multiple tasks share the environment dynamics, it would be more effective for a robot to simultaneously learn the tasks in the same framework. 

\begin{figure}[t]
  \begin{center}
      \includegraphics[width=1.0\columnwidth]{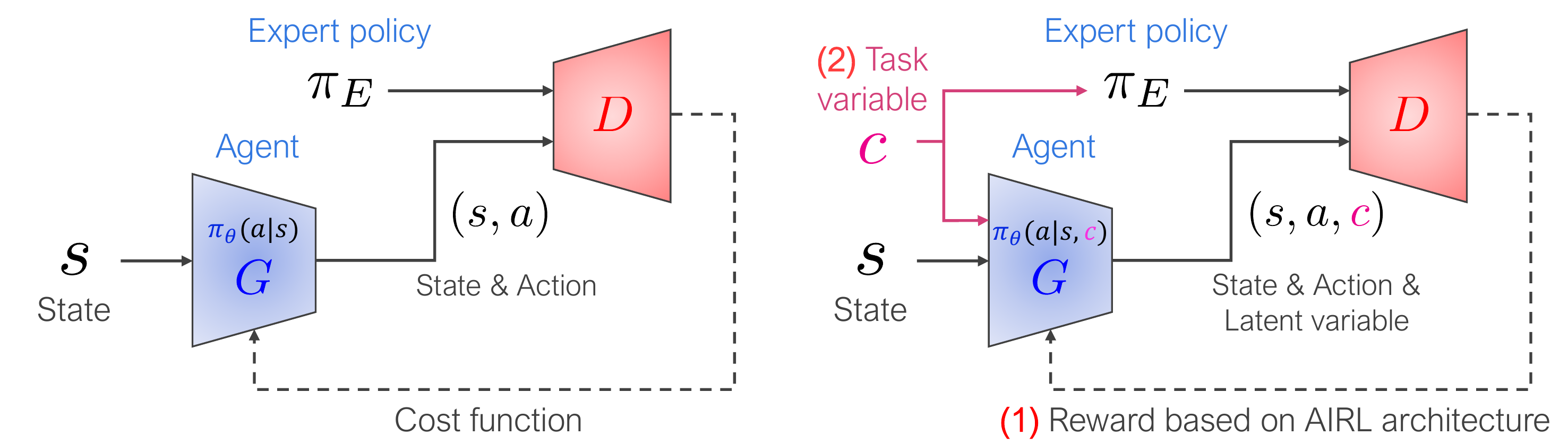}
      \caption{Architectures of GAIL (left) and S-GAIL (right). (1) and (2) indicate our ideas for extending GAIL.}
      \label{fig:proposed_method}
  \end{center}
\end{figure}

This study proposes a new framework called situated GAIL (S-GAIL) that extends GAIL, AIRL, and InfoGAIL to learn multiple reward functions and multiple policies in a single framework.
Figure \ref{fig:proposed_method} shows the differences between GAIL (left) and S-GAIL (right).
Our key contributions are twofold:
1) to employ the discriminator's structure proposed in AIRL (as shown (1) in Figure \ref{fig:proposed_method}) and
2) to introduce a task variable used in InfoGAIL to both the discriminator and generator (as shown (2) in Figure \ref{fig:proposed_method}).
The first concept enables our framework to directly estimate reward functions.
As suggested in \cite{Fu_AIRL2017}, recovering reward functions are expected to improve the generalization capabilities of acquired policies.
The second concept enables our framework to deal with multiple rewards as well as multiple policies. 
In contrast to InfoGAIL, which uses a latent variable only in the generator, the task variable in S-GAIL conditions the discriminator and generator to differentiate between different goals and means.
In addition to the above key concepts, we adopt a technique to improve the learning speed and task performance.
It is known that entropy regularization can avoid over-fitting during learning \cite{Haarnoja2017,Nachum2017}.
We additionally adjust the coefficient for regularization from a smaller to a larger value so that S-GAIL first replicates an expert's behaviors precisely and then optimizes policies using its own acquired dynamics.

The rest of this article is organized as follows. Section 2 presents previous work related to imitation learning.
GAIL and InfoGAIL, which are the bases of the proposed framework, are explained in more detail.
Sections 3 and 4 describe the architecture of the proposed model after providing preliminaries.
Two experiments using a discrete grid world and a continuous robot arm model are presented in Section 5.
Finally, Section 6 provides conclusions and future issues to be addressed.


\section{Related work}
\label{sec:citations}
GAIL was proposed as a framework for imitation learning \cite{Ho_GAIL2016}.
The goal of imitation learning is to replicate an expert's behaviors without any a priori knowledge about the task or environment. 
As in a generative adversarial framework, IRL and RL in GAIL are formulated as competitive optimization problems.
The discriminator and generator compete to simultaneously learn a cost function and policy.
That is, the generator learns to produce behaviors similar to those presented by an expert, while the discriminator learns to discriminate the output of the generator from the expert's behaviors.
This competitive learning framework based on the architecture of GAN ensures that it has a unique optimal cost function and policy.

However, Fu et al. \cite{Fu_AIRL2017} have claimed that the discriminator of GAIL cannot recover a reward function behind an expert's behaviors.
GAIL learns a policy to replicate the behaviors without explicitly addressing an IRL problem.
It is expected that reward functions help a learner acquire robust behaviors. 
Rewards can disentangle learned policies from changes in the environmental dynamics.
AIRL \cite{Fu_AIRL2017} was thus proposed to cope with this problem by placing a specific form on the discriminator to derive a reward function (see Section 4 for more detail). 

Researchers have been also extending GAIL to apply it to more challenging problems.
For example, InfoGAIL \cite{Li_infoGAIL2017} was proposed to learn multiple behaviors to achieve a goal.
It was assumed in the original GAIL that an expert follows a single optimal policy to perform a task.
If the expert has multiple policies (e.g., to reach a goal position from the left and right sides while avoiding an obstacle at the center), GAIL fails to separately represent them and instead acquires the mean of the multiple policies (e.g., to reach the goal from the center, which is not achievable owing to the obstacle).
InfoGAIL solves this problem by introducing latent variables to the generator.
It employs a regularization term to maximize the mutual information between the latent variables and the output trajectories of the generator.
In this way, InfoGAIL learns to acquire multiple policies conditioned by the latent variables.
However, applying InfoGAIL to multitask imitation learning remains difficult.
InfoGAIL assumes that a single reward function governs an expert's behaviors.
In other words, if an expert has multiple goals represented by multiple reward functions (e.g., reaching two distinct goal positions), InfoGAIL cannot estimate them.
Because the latent variable is used only for the generator, InfoGAIL cannot differentiate the reward functions for different tasks.

In contrast, another extension of GAIL, called conditional GAIL \cite{Merel2017}, has been proposed by introducing a latent variable to both the discriminator and generator.
Although its capability has not been demonstrated, we consider that this framework can represent multiple reward functions.
However, it is unclear if conditional GAIL has a unique optimal solution for its objective function in the adversarial framework.
OptionGAN \cite{Henderson2018}, which includes policy options to the GAIL framework, formulates a method for learning joint reward policy options with adversarial methods in IRL.
The policy option is a type of sub-goal in the RL framework and is used with other policies to represent a complex and high order policy based on the concept of a mixture of experts.
OptionGAN divides the generator into policy options and employs multiple discriminators as a mixture of experts of reward functions for the generators. 
However, the discriminator of conditional GAIL and OptionGAN cannot recover reward functions as well as InfoGAIL because they do not specify the discriminator's structure as in AIRL.
The goal of this study is to propose a new framework to address these issues.

\section{Preliminaries}
\label{sec:Preliminaries}
Let the tuple $(\mathcal{S}, \mathcal{A}, P, \mathcal{R}, \gamma, \rho_0, T)$ be a finite-horizon Markov decision process (MDP), where $\mathcal{S}$ and $\mathcal{A}$ are the state and action spaces respectively, and $P: \mathcal{S} \times \mathcal{A} \times \mathcal{S} \to \mathbb{R}$ is the state transition probability of the system dynamics.
At a discrete time $t$, an agent observes a state $s_t$ and selects an action $a_t$ according to the agent's policy $\pi(a_t|s_t) = P(a_t|s_t)$, and it receives an immediate reward $r(s_t,a_t) \in \mathcal{R}$ from the environment.

The goal of the RL problem is to acquire a policy to maximize the sum of the expectation of the $\gamma$-discounted reward, where $\gamma \in [0,1)$ is a discount factor.
To consider the entropy--regularized MDP (ER-MDP), an entropy regularized term is added to the objective function $\eta(\pi)=\mathbb{E}_{\pi} \bigl[ \sum_{t=0}^T \gamma r(s_t,a_t) | s_0=s, a_0=a \bigr]$:
\begin{align}
{\rm RL} = \pi(a|s) &\in \underset{\pi}{\rm argmax} \,\,\, \eta(\pi) \nonumber \\
&\in \underset{\pi}{\rm argmax} \,\,\, \mathbb{E}_{\pi} \Biggl[ \sum_{t=0}^T \gamma^{t} \bigl( r(s_t,a_t) - \omega H(\pi(a_t|s_t)) \bigr) |s_0=s, a_0=a \Biggr], \nonumber
\end{align}
where $H(\pi(a|s)) \triangleq \mathbb{E}_{\pi} \bigl[ - \log \pi(a|s) \bigr]$ is the entropy of policy $\pi$, and $\omega$ is the weight of the entropy regularization term.
The value function $V^{\pi}$ and action value function defined in the ER-MDP satisfy the following Bellman equation:
\begin{align}
V^{\pi}(s) &= \sum_{a} \pi(a|s) \biggl\{ \mathcal{R}(s,a) - \omega \log \pi(a|s) + \gamma \sum_{s'} P(s'|s,a) V^{\pi}(s') \biggr\}, \nonumber \\
Q^{\pi}(s,a) &= \mathcal{R}(s,a) + \gamma \sum_{s'} P(s'|s,a) V^{\pi}(s'). \nonumber
\end{align}
The optimal policy $\pi^*$ in the ER-MDP follows the probability distribution given by the following equation with the optimum value function $V^*$:
\begin{align}
\pi^*(a|s) = \frac{\exp \bigl\{ \bigl( \mathcal{R}(s,a) + \gamma \sum_{s'} P(s'|s,a) V^{*}(s')\bigl) / \omega \bigr\}}{\exp (V^*(s)/ \omega)}. \nonumber
\end{align}
Then, the relationship between the optimal policy $\pi^*$ and the advantage function has been held as follows:
\begin{align}
A^{*}(s,a) &= \mathcal{R}(s,a) + \gamma \sum_{s'} P(s'|s,a) V^{*}(s') - V^{*}(s) \nonumber \\
&= Q^{*}(s,a) - V^{*}(s) = \omega \log \pi^*(a|s). \nonumber
\end{align}

Assuming that the expert's behavior follows RL principles, imitation learning is used to estimate the reward function from the expert's behavior.
The expert's behavior data are given as a set of trajectories $\tau$ in which state and action pairs are arranged in chronological order for each episode.
IRL estimates the reward function by solving the following optimization problem \cite{Ziebart2008}:
\begin{align}
{\rm IRL} = \underset{l}{\rm maximize} \,\, \biggl( \min_{\pi} \,\, - H(\pi(a|s)) + \mathbb{E}_{\pi} \bigl[l(s,a) \bigr] \biggr) - \mathbb{E}_{\pi_E} \bigl[ l(s,a) \bigr]. \nonumber
\end{align}


\section{S-GAIL: A proposed framework for estimating multiple rewards and policies}
\label{sec:Situated GAIL}
This section presents a detailed formulation of imitation learning based on GAIL.
First, we introduce the existing models, GAIL, InfoGAIL, and AIRL as components of the proposed model.
Then, the proposed model S-GAIL is presented.

\subsection{Basic components of S-GAIL}
\subsubsection{Reward estimation based on an adversarial training: GAIL and AIRL}
Ho and Ermon \cite{Ho_GAIL2016} showed that the synthesis problem of IRL and RL can be written as the following optimization problem:
\begin{align}
{\rm RL} \circ {\rm IRL} = \underset{\pi \in \Pi}{\rm argmin} \,\,\, - H(\pi(a|s)) + \psi^*(\rho_{\pi}(s,a) - \rho_E(s,a)), \label{eq:GAIL}
\end{align}
where $H(\pi)$ is the entropy and $\rho_{\pi}(s, a)$ is the joint distribution of state $s$ and action $a$ under policy $\pi$.
When we consider a set of $\rho_{\pi}(s, a)$ that satisfies the Bellman constraint $\mathcal{M} = \{ \rho_{\pi} : \pi \in \Pi \} = \bigl\{ \rho : \rho \geq 0 \,\, {\rm and} \,\, \sum_{a} \rho(s, a) = \rho_0(s) + \gamma \sum_{s',a} P(s|s', a)\rho(s, a) \bigr\}$, it is shown that $\rho$ satisfies $\rho \in \mathcal{M}$, which corresponds to policy $\pi$ on a one-on-one basis, and the relational expression of $\pi(a|s) = \rho_{\pi}(s, a) / \sum_{a'}\rho_{\pi}(s, a')$ holds~\cite{Syed2008}.
$\psi^*:\mathbb{R}^{\mathcal{S} \times \mathcal{A}} \to \bar{\mathbb{R}}$ is a conjugate function of the convex regularization function $\psi(l):\mathbb{R}^{\mathcal{S} \times \mathcal{A}} \to \bar{\mathbb{R}}$ with the cost function $l(s, a)$ (having the opposite sign of the reward function) as a variable and it satisfies the following equation:
\begin{align}
\psi^*(\rho_{\pi}(s,a) - \rho_E(s,a)) = \underset{l \in \mathbb{R}^{\mathcal{S} \times \mathcal{A}}}{\rm sup}\, \sum_{s,a} \bigl( \rho(s,a) - \rho_E(s,a) \bigr)l(s,a) - \psi(l), \nonumber
\end{align}
where $\bar{\mathbb{R}}$ is an extended real number.

The objective function in equation (\ref{eq:GAIL}) has a saddle point when it is considered as a function with $\rho(s,a)$ and $l(s,a)$ as its variables.
Additionally, it is guaranteed that the saddle point is the only optimal solution of the objective function.
GAIL considers this optimization problem as the learning of a discriminator and a generator.
The learning rule of GAN can then be applied \cite{Goodfellow2014}:
\begin{align}
\underset{\bm w}{\rm maximize} \,\,\, &\mathbb{E}_{\pi_E} \bigl[ \log \bigl( D_{\bm w}(s, a) \bigr) \bigr] + \mathbb{E}_{\pi_{\rm \theta}} \bigl[ \log \bigl( 1 - D_{\bm w}(s, a) \bigr) \bigr] \nonumber \\
\underset{\bm \theta}{\rm minimize} \,\,\, &\mathbb{E}_{\pi_{\bm \theta}} \bigl[ \log \bigl( 1 - D_{\bm w}(s, a) \bigr) \bigr] - \lambda H(\pi_{\bm \theta}), \nonumber
\end{align}
where ${\bm w}$ and ${\bm \theta}$ are the discriminator and generator parameters, respectively, and $\lambda$ is the hyperparameter for the entropy term.
$D(\cdot)$ is the output of the discriminator; it indicates the probability that the input state $s$ and action $a$ are those of the expert.
The discriminator learns to correctly identify whether the distribution that generated the state--action pair is a generator or an expert.
The generator learns to output the selection probability of the action so that the discriminator confuses the generator's trajectories with those of the expert.

It is expected that the policy and reward function learned through RL and IRL can be generalized to unknown states and actions.
However, it has been indicated that the discriminator of GAIL cannot recover the reward function from expert demonstrations because of the unrestricted structure of the discriminator.
Some researchers have discussed the equivalence between RL, IRL, and GANs \cite{Finn_CB2016, Fu_AIRL2017, Pfau2016}.
Fu et al. \cite{Fu_AIRL2017} claimed that the discriminator of GAIL is unsuitable for recovering the reward function.
To represent reward functions, AIRL employs a special structure for the discriminator corresponding to an odds ratio between the policy and the exponential reward according to \cite{Finn_CB2016}:
\begin{align}
D(s,a) = \frac{\exp(f(s,a))}{\exp(f(s,a))+\pi(a|s)}, \label{eq:AIRL}
\end{align}
where $ f(s, a) $ is an arbitrary function (e.g., neural network) and $ \pi(a|s) $ is the probability of action $a$ with state $s$.
The policy is trained to maximize $\log(1 - D) - \log D$.
According to this formulation, the function $f$ becomes an advantage function $A$ of the optimal policy $\pi^*$, $f^*(s, a) = \log \pi^*(a|s) = A^*(s, a)$.
This odds ratio structure enables the discriminator to estimate the reward function from the expert demonstration in adversarial training.

\subsubsection{Modeling multiple policies by latent variables: InfoGAIL}
GAIL and AIRL assume that the expert's behaviors follow a single policy; that is, they cannot represent multimodal trajectories, e.g., the expert reaches a goal from both the left and right sides.
To overcome this limitation, several extensions \cite{Li_infoGAIL2017, Hausman2017, Lin_ACGAIL2017, Merel2017} based on the idea of InfoGAN \cite{Chen_infoGAN2016} have been proposed, which introduces latent variables to the generator to represent multimodal distributions.
InfoGAIL \cite{Li_infoGAIL2017} was proposed to infer the latent structure of expert behaviors in an unsupervised manner.
Further, a similar idea to InfoGAIL has been proposed in \cite{Hausman2017}.
Both models introduce latent variable $c$ to represent multiple policies in the generator and maximize the mutual information between $c$ and the trajectory $\tau = \{s_0, a_0, \dots, s_T, a_T \} \sim \pi_{\bm \theta}$.
Then, the policy $\pi_{\bm \theta}$ is selected from the mixture of policies through $p(\pi_{\bm \theta}|c)$; that is, the trajectories $\tau$ are generated by the conditional policy $\pi_{\bm \theta}(a|s,c)$. 
The mutual information $ I (c; \tau) $ is similarly expressed as \cite{Chen_infoGAN2016}
\begin{align}:
I(c; \tau \sim \pi_{\bm \theta}) &= H(c) - H(c | \tau \sim \pi_{\bm \theta}) \nonumber \\
&= \mathbb{E}_{c \sim p(c), (s,a) \sim \pi_{\bm \theta}} \biggl[ \mathbb{E}_{c' \sim P(c|s,a)} \bigl[ \log P(c'|s,a) \bigr] \biggr] + H(c) \nonumber \\
&= \mathbb{E}_{c \sim p(c), (s,a) \sim \pi_{\bm \theta}} \biggl[ D_{KL} \bigl( P(\cdot|s,a) || Q(\cdot|s,a) \bigr) + \mathbb{E}_{c' \sim P(c|s,a)} \bigl[ \log Q(c'|s,a) \bigr] \biggr] + H(c) \nonumber \\
&\geq \mathbb{E}_{c \sim p(c), (s,a) \sim \pi_{\bm \theta}} \biggl[ \mathbb{E}_{c' \sim p(c)} \bigl[ \log Q(c'|s,a) \bigr] \biggr] + H(c) \nonumber \\
&= \mathbb{E}_{c \sim p(c), (s,a) \sim \pi_{\bm \theta}} \bigl[ \log Q(c|s,a) \bigr] + H(c). \nonumber
\end{align}
In order to maximize the lower bound of the mutual information $ I (c; \tau) $, the auxiliary distribution $Q(c|s,a)$, which can be provided by a neural network, is trained.
In the end, the objective functions of the discriminator and generator of InfoGAIL are
\begin{align}
\underset{\bm w}{\rm maximize} \,\,\, &\mathbb{E}_{\pi_E} \bigl[ \log D_{\bm w}(s,a) \bigr] + \mathbb{E}_{c \sim p(c), \, \pi_{\rm \theta}} \bigl[ \log \bigl( 1- D_{\bm w}(s,a) \bigr) \bigr], \nonumber \\
\underset{\bm \theta}{\rm minimize} \,\,\, &\mathbb{E}_{c \sim p(c), \, \pi_{\bm \theta}} \bigl[ \log \bigl( 1 - D_{\bm w}(s,a) \bigr) \bigr] - \lambda_1 H(\pi_{\bm \theta}) - \lambda_2 \mathbb{E}_{c \sim p(c), (s,a) \sim \pi_{\bm \theta}} \bigl[ \log Q(c|s,a) \bigr]. 　\nonumber
\end{align}

According to the above objective functions, InfoGAIL can infer the latent information in the expert's demonstration and learn multiple policies from the demonstrations by the generator.
However, there is still a limitation: InfoGAIL cannot represent multiple reward functions.
Because the latent variable was only introduced to the generator, the discriminator cannot differentiate the reward for multiple tasks.

\subsection{Proposed model: S-GAIL}
We propose S-GAIL that achieves robust multitask learning within a single framework.
S-GAIL integrates two mechanisms into GAIL so as to take advantage of InfoGAIL and AIRL.
It 1) employs a task variable in both the generator and discriminator and 2) introduces a specific structure in the discriminator to recover reward functions.

First, task variables are employed for both the discriminator and generator to estimate the reward functions corresponding to different tasks to acquire policies corresponding to each task.
Figure \ref{fig:proposed_method} shows a schematic of the proposed method.
We assume that the dataset of the expert's behavior includes the task variable $c$, which differentiates multiple tasks.
The objective functions of S-GAIL are
\begin{align}
\underset{\bm w}{\rm maximize} \,\,\, & \mathbb{E}_{c \sim p(c), \, \pi_{E}} \bigl[ \log \bigl(D_{\bm w}(s,a,c) \bigr) \bigr] +  \mathbb{E}_{c \sim p(c), \, \pi_{\theta}} \bigl[ \log \bigl(1 - D_{\bm w}(s,a,c) \bigr) \bigr], \label{eq:proposed_Discriminator_loss} \\
\underset{\bm \theta}{\rm minimize} \,\,\, & \mathbb{E}_{c \sim p(c), \, \pi_{\theta}} \bigl[ \log \bigl(1 - D_{\bm w}(s,a,c) \bigr) \bigr] - \mathbb{E}_{c \sim p(c), \, \pi_{\theta}} \bigl[ \log \bigl(D_{\bm w}(s,a,c) \bigr) \bigr] \label{eq:proposed_Generator_loss}
\end{align}

Then, the odds ratio structure of the discriminator is adopted from AIRL \cite{Fu_AIRL2017} to estimate reward functions.
It is given as follows:
\begin{align}
D_{\bm w}(s,a,c) = \frac{\exp \bigl( f_{\bm w}(s,a,c) \bigr)}{\exp \bigl( f_{\bm w}(s,a,c) \bigr) + \pi_{\theta}(a|s,c)}, \label{eq:D_sGAIL}
\end{align}
where $f$ is arbitrary function of the state $s$, the action $a$, and the kind of task $c$.
If the above problems converge, $f^*(s,a,c) = \log \pi^*(a|s,c) = Q^*(s,a,c) - V^*(s,c) = A^*(s,a,c)$ is satisfied, as in a GAN \cite{Goodfellow2014}.
$\pi^*(a|s,c)$ is the optimal policy, and $V^*(s,c)$ and $Q^*(s,a,c)$ are the optimal value and action value function, respectively.
$A^*(s,a,c)$ is the advantage function that follows $\pi^*(a|s,c)$.
The second equality holds when the regularization coefficient in an ER-MDP is $1$ \cite{Nachum2017}.

The learning rule of S-GAIL has an optimum solution, similar to a GAIL formulation; in other words, the following proposition holds.

\begin{prop}
The solution of the synthesis problem of IRL and RL in the process of entropy regularization by the introduction of task variables is equivalent to the solution of the following optimization problem:
\begin{align}
\underset{\pi \in \Pi}{\rm argmin} \,\,\, - H(\pi(a|s,c)) + \psi^*(\rho(s,a,c) - \rho_E(s,a,c)), \label{eq:IRL_RL_sGAIL}
\end{align}
and it has a saddle point.
\end{prop}

Proposition 1 can be proved by considering the RL and IRL problem under the ER-MDP that introduces the task variable and by showing that their objective functions can be reduced to equation (\ref{eq:IRL_RL_sGAIL}), as in GAIL.
However, unlike in GAIL, the simultaneous distribution $\rho(s,a,c)$ is extended to the task space, and $H(\pi) = \mathbb{E}_{\pi, c}[ - \log \pi(a|s,c)]$ is considered.

\begin{prop}
The solution of the synthesis optimization problems (\ref{eq:proposed_Discriminator_loss}) and (\ref{eq:proposed_Generator_loss}) is equivalent to the solution of the optimization problem (\ref{eq:IRL_RL_sGAIL}).
\end{prop}

Proposition 2 can be proved by substituting the structure of discriminator in equations (\ref{eq:proposed_Discriminator_loss}) and  (\ref{eq:proposed_Generator_loss}).
They correspond to IRL and RL.
Propositions 1 and 2 confirm that the proposed method has only one optimal solution.

\subsubsection{Introduction of a coefficient for the entropy-regularized term and its adjustment during training}
The objective function of the generator of S-GAIL (equation (\ref{eq:proposed_Generator_loss})) can be separated into an accuracy of imitation term $f_{\omega}(s,a,c)$ and an entropy-regularized term as follows:
\begin{align}
&\mathbb{E}_{c \sim p(c), \, \pi_{\theta}} \bigl[ \log \bigl(1 - D_{\bm w}(s,a,c) \bigr) \bigr] - \mathbb{E}_{c \sim p(c), \, \pi_{\theta}} \bigl[ \log D_{\bm w}(s,a,c) \bigr] \nonumber \\
&= - \mathbb{E}_{c \sim p(c), \, \pi_{\theta}} [f_{\bm w}(s,a,c)] + \mathbb{E}_{c \sim p(c), \, \pi_{\theta}} [\log \pi_{\theta}(a|s,c)] \nonumber \\
&= - \mathbb{E}_{c \sim p(c), \, \pi_{\theta}} [f_{\bm w}(s,a,c)] - H(\pi). \nonumber
\end{align}
It is known that the maximization of the entropy of the policy in the objective function of RL leads to the acquisition of a unique optimal solution.

However, in an adversarial learning manner, a policy that has a high entropy is easy to discriminate from the expert policy.
In other words, S-GAIL possibly fails because the discriminator becomes stronger than the generator in the early stage of training.
To avoid this problem, we introduce a coefficient to correct the entropy-regularized term to the generator's objective function to adjust the ratio of the entropy-regularized term:
\small
\begin{align}
&\underset{\bm \theta}{\rm minimize} \,\,\, - \mathbb{E}_{c \sim p(c), \, \pi_{\theta}} [f_{\bm w}(s,a,c)] + \mathbb{E}_{c \sim p(c), \, \pi_{\theta}} [\log \pi_{\theta}(a|s,c)] - \beta \mathbb{E}_{c \sim p(c), \, \pi_{\theta}} \bigl[ \log \bigl( \pi_{\theta}(a|s,c) \bigr) \bigr], \nonumber \\
&= \underset{\bm \theta}{\rm minimize} \,\,\,  - \mathbb{E}_{c \sim p(c), \, \pi_{\theta}} \bigl[ f_{\bm w}(s,a,c) \bigr] + (1 - \beta) \mathbb{E}_{c \sim p(c), \, \pi_{\theta}} \bigl[ \log \bigl( \pi_{\theta}(a|s,c) \bigr) \bigr]
\end{align}
\normalsize
By introducing the parameter $\beta$, we can modulate the balance of both terms.
At the beginning of learning, we reduce the effect of the entropy (i.e., $\beta$ is set to a large value) to approximate the expert's policy.
Then, we set $\beta$ to a small value to gain the effect of the entropy term for agent exploration to learn a robust policy in the later stage of learning. 

Algorithm \ref{alg:algorithm1} shows the calculation procedure for S-GAIL.
Any approximator can be used as the discriminator and generator when it is differentiable.
Additionally, we prepared a value function $V(s,c)$, which was parameterized by a neural network to estimate the advantage function.
To optimize the parameters of the generator, trust region policy optimization \cite{Schulman_TRPO2015} is used with the advantage function, and the discriminator and value function are updated using the Adam optimizer \cite{Kingma2014}.

\begin{algorithm}[t]
\caption{Situated GAIL} 
\label{alg:algorithm1}
\begin{algorithmic}
\STATE{\textbf{Input:}}
\STATE{Expert trajectories, task variables: $\tau_E \sim \pi_E, \,\, c_E \sim p(c)$}
\STATE{Initial parameters of the generator, discriminator, and value function: ${\bm \theta}={\bm \theta}_{0}, \,\, {\bm w}={\bm w}_{0}, \,\, {\bm \phi}={\bm \phi}_{0}$}
\STATE{Entropy-regularized correction parameter and its scheduling parameter: $\beta, \,\, \Delta \beta$}
\FOR{$i = 0,1,2,...$}
\STATE{Sample task variables and trajectories:}
\STATE{~~~~~~~~~~$c_i \sim p(c), \,\, \tau_i \sim \pi_{\theta_i}$}
\STATE{Discriminator update:}
\STATE{~~~~~~~~~~${\bm w}_{i+1} = {\bm w}_{i} + \alpha_{\bm w} \Delta {\bm w}_{i}$}
\STATE{~~~~~~~~~~${\rm where} \,\,\, \Delta {\bm w}_{i} = \mathbb{E}_{c,\pi_E} \bigl[ \nabla_{\bm w} \log \bigl( D_{\bm w}(s,a,c) \bigr) \bigr] + \mathbb{E}_{c,\pi_{\bm \theta}} \bigl[ \nabla_{\bm w} \log \bigl(1 - D_{\bm w}(s,a,c) \bigr) \bigr] $}
\STATE{Value function update:}
\STATE{~~~~~~~~~~${\bm \phi}_{i+1} = {\bm \phi}_{i} + \alpha_{\bm \phi} \Delta {\bm \phi}_{i}$}, 
\STATE{~~~~~~~~~~${\rm where} \,\,\,\Delta {\bm \phi}_{i} = \nabla_{\bm \phi} \mathbb{E}_{c,\pi_{\bm \theta}} \bigl[ \bigl( \mathcal{R}_{\pi_{\bm \theta}} - V_{\phi_i}(s,c) \bigr)^2 \bigr] $},
\STATE{~~~~~~~~~~~~~~$\,\,\,\,\,\,\,\,\,\,\,\,\, \mathcal{R}_{\pi_{\bm \theta}} = \log \bigl(D_{{\bm w}_{i+1}}(s,a,c) \bigr) - \log \bigl(1 - D_{{\bm w}_{i+1}}(s,a,c) \bigr) + \beta \log \bigl( \pi_{\theta}(a|s,c) \bigr) $}
\STATE{Generator update:}
\STATE{~~~~~~~~~~ Using policy gradient algorithm with the following advantage function}
\STATE{~~~~~~~~~~$\mathcal{A}_{\pi_{\bm \theta}} = \mathcal{R}_{\pi_{\bm \theta}} + \gamma V_{\bm \phi_i}(s',c) - V_{\bm \phi_i}(s,c)$}
\STATE{Modulate entropy correction parameter:}
\STATE{~~~~~~~~~~$\beta \leftarrow \beta + \Delta \beta$}
\ENDFOR

\end{algorithmic}
\end{algorithm}

\section{Experiments and results}
\label{sec:result}
We conducted two experiments to demonstrate the advantages of S-GAIL over existing methods.
The first experiment employed a simple grid world, which allowed us to closely analyze the internal representation acquired by S-GAIL.
The second experiment used a robot arm simulator working in a continuous space to demonstrate the scalability of S-GAIL. 

\subsection{Grid world}
The first experiment was designed as a maze within a grid world, where an agent had to imitate an expert reaching multiple target locations.
Namely, the agent should acquire different actions in the same state corresponding to the task reward functions.
First, we compared the task performance of S-GAIL and existing methods while visualizing their value functions.
Then, the characteristics of S-GAIL (i.e., the effect of the coefficient of the entropy-regularized term and the advantage of representing multiple rewards in a single network) were evaluated.

\subsubsection{Setting}
\begin{figure}[htbp]
  \begin{center}
      \includegraphics[width=0.4\columnwidth]{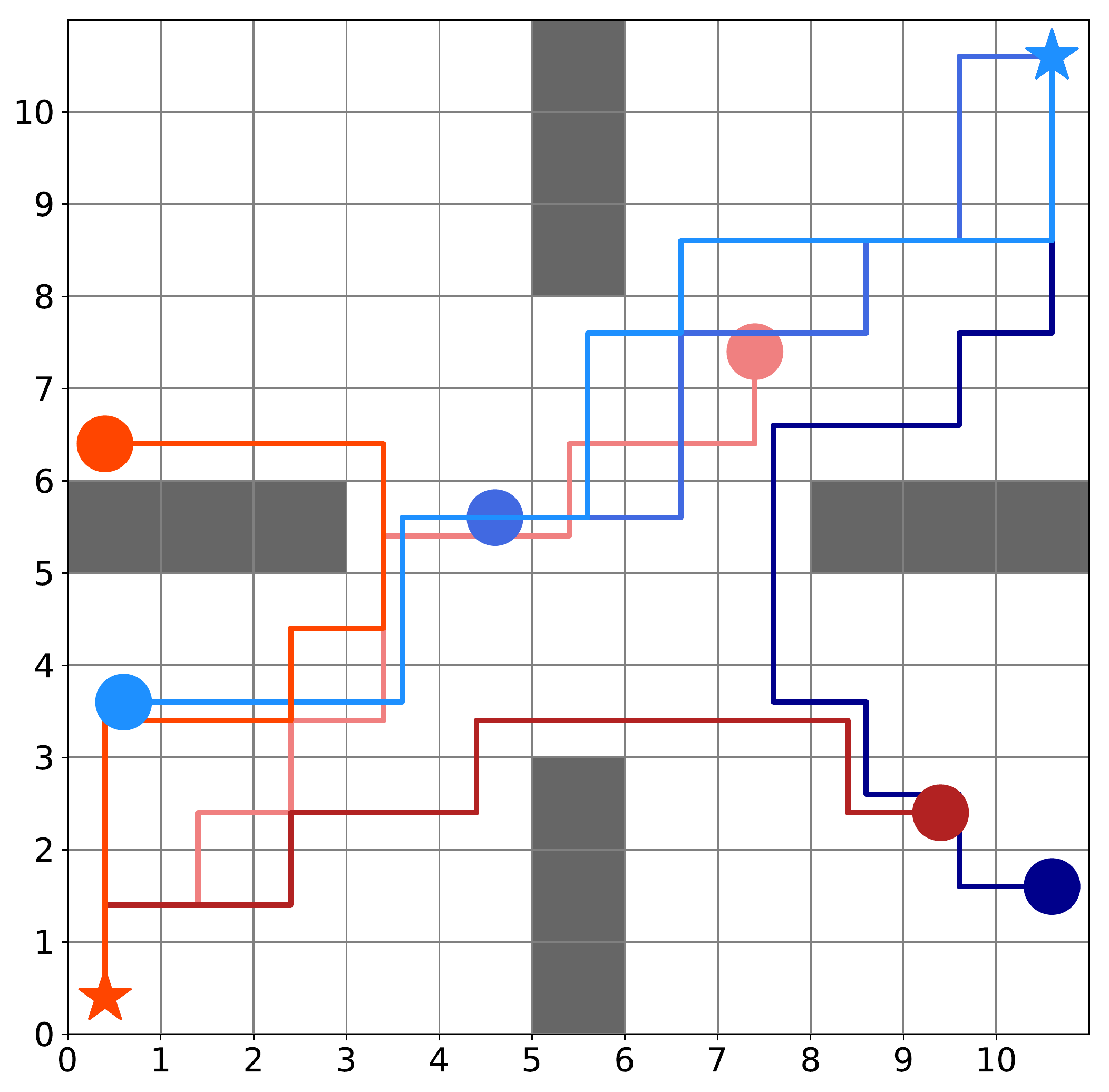}
      \caption{Grid world and examples of expert trajectories. $\bullet$ denotes an initial state and $\star$ denotes a goal state.}
      \label{fig:gridworld}
  \end{center}
\end{figure}

Figure \ref{fig:gridworld} shows an $11 \times 11$ grid world used in the experiment.
The state of the grid world was represented by $(x,y)$, where $x,y \in [0,10]$; the action was a four-dimensional one-hot vector representing the direction of movement (i.e., ${right, up, left, down}$).
The state transition of the agent was deterministic.
The agent could not move to the puddle states indicated in black.
The task variable $c$ was represented by a one-hot vector, i.e., a discrete variable.

In the current experiment, we defined two tasks to reach different goal locations: (0, 0) and (10, 10), indicated by the red $\star$ and blue $\star$ in Figure \ref{fig:gridworld}.
Each task was named task 1 and task 2, respectively.
For each task, the expert's behaviors were sampled 30 times using random initial positions denoted by the $\bullet$ symbol.
The expert always took the shortest path to reach the target locations.
The task variables $c$ for each task were represented by $c_1 = (1, 0, 0)$ and $c_2 = (0, 1, 0)$.
They were given with the corresponding expert's behavior.

\begin{figure}
    \centering
    \subfigure[Generator]{\includegraphics[width=0.25\columnwidth]{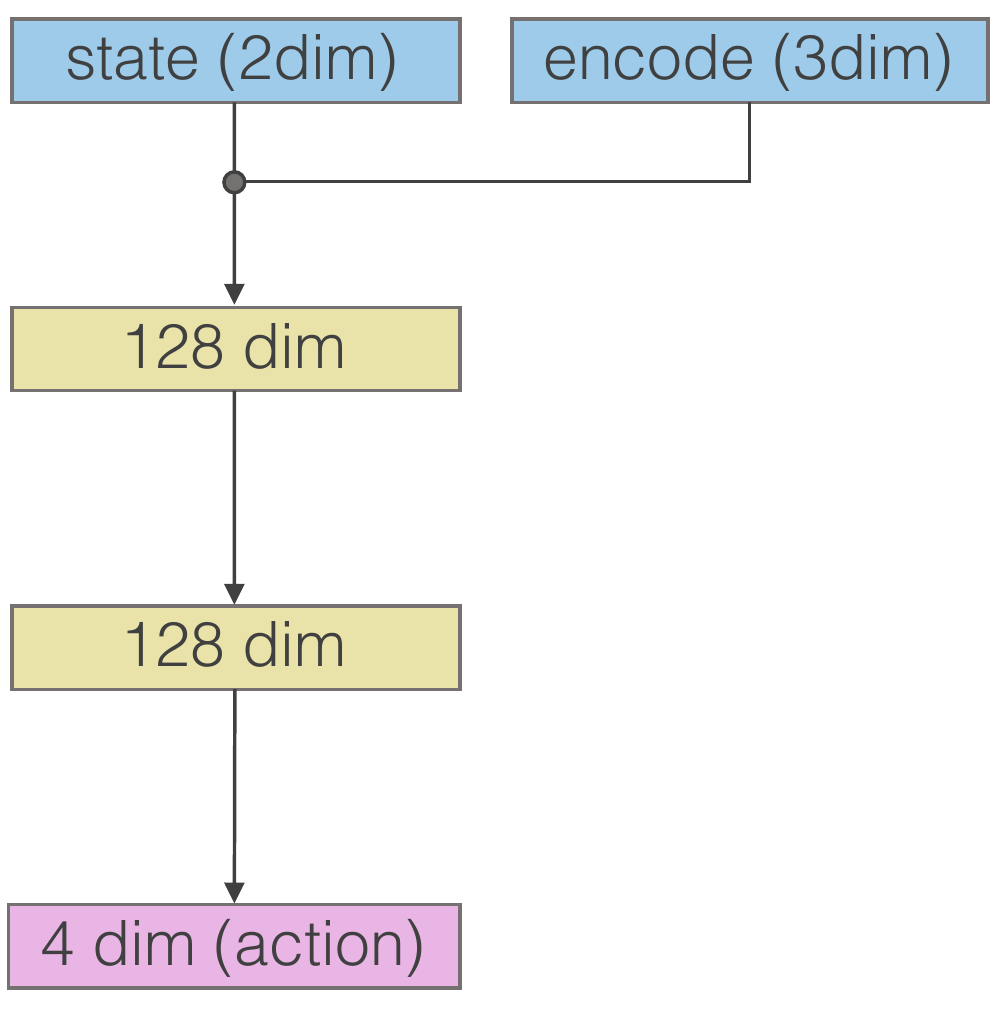}}
    \subfigure[Discriminator]{\includegraphics[width=0.4\columnwidth]{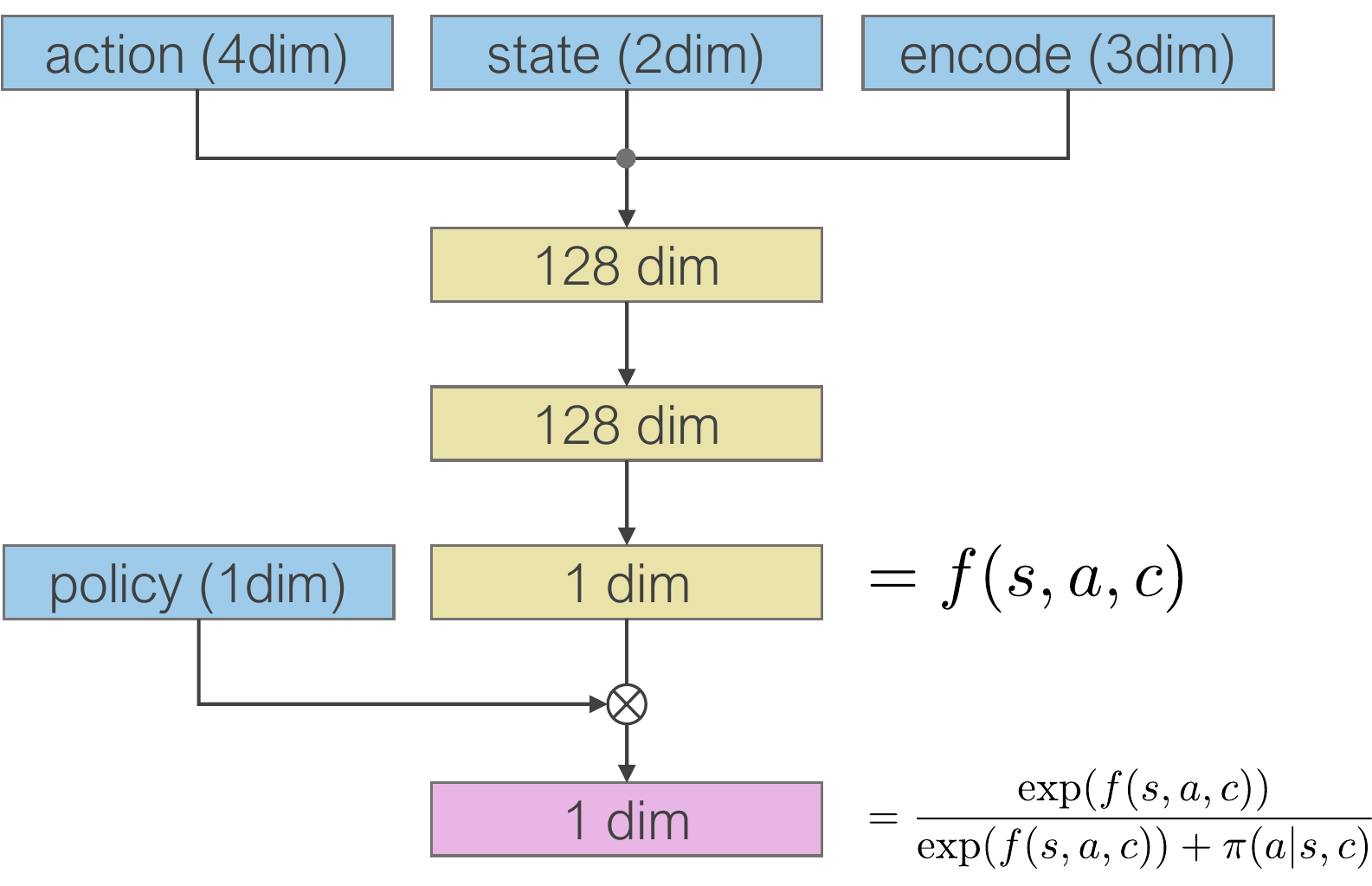}}
    \subfigure[Value function]{\includegraphics[width=0.25\columnwidth]{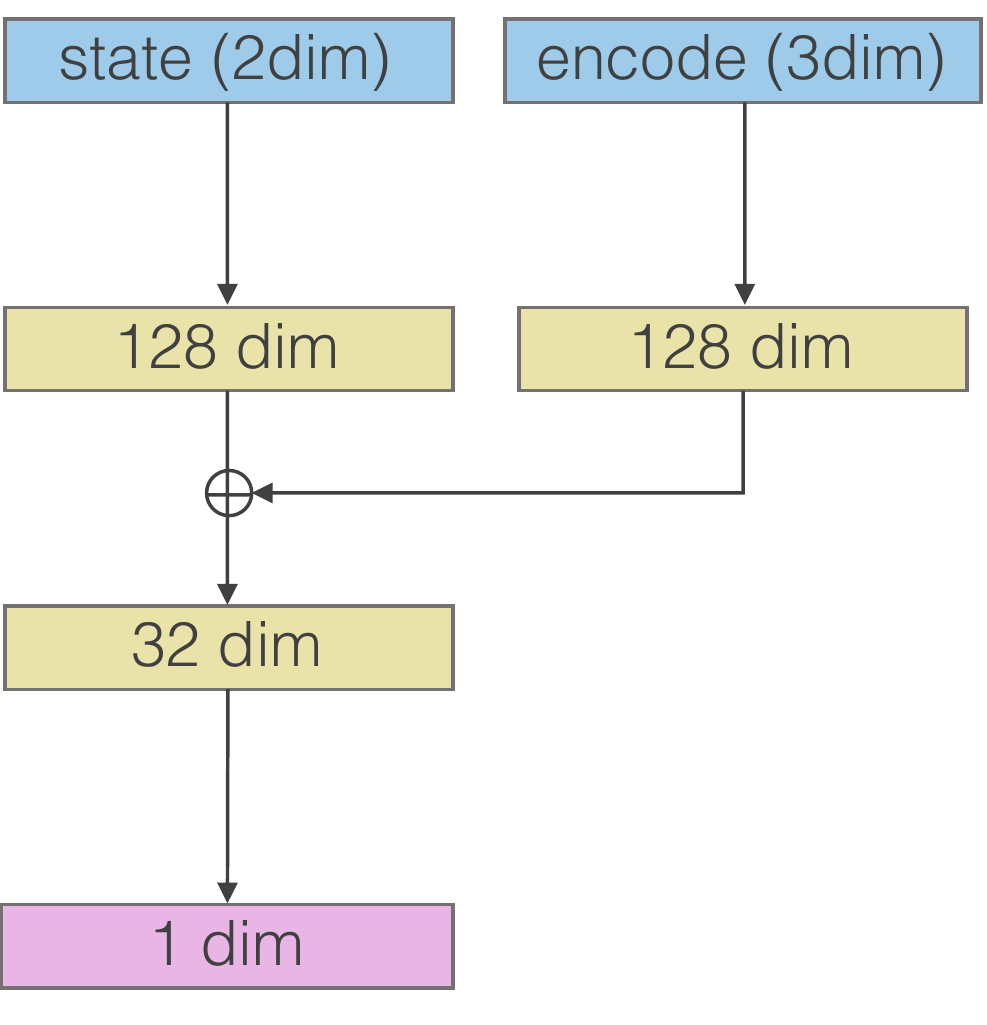}}
    \caption{Structures of the generator, discriminator, and value function networks. The numbers in boxes indicate the number of nodes in each layer. $\bullet$ means the concatenate operation. $\oplus$ denotes the add operation in each node. $\otimes$ transforms the function $f$ and the policy into the discriminator of the AIRL structure (equation \ref{eq:D_sGAIL}).}
    \label{fig:Network}
\end{figure}

In this experiment, the generator, discriminator, and value function were parameterized by neural networks.
Figure \ref{fig:Network} illustrates the network structures.
The generator network had five input nodes: two nodes corresponding to the agent's state and three nodes corresponding to the task variable $c$, and four output nodes corresponding to the agent's action.
The discriminator network had nine input nodes for the agent's action, state, and task variable.
Additionally, the output of the discriminator's hidden layer (i.e., the function $f(s,a,c)$) was combined with the policy from the generator to construct the odds ratio structure of AIRL's discriminator (such as equation \ref{eq:D_sGAIL}) \cite{Fu_AIRL2017}.
This formulation can represent multiple reward functions corresponding to task variable $c$.
The value function network had five inputs for the agent's state and task variable.
All networks used the leaky relu node as their activation function, except the output layer of the generator, which used the softmax function to represent the probabilities of actions.

  


\subsubsection{Result 1: Comparison of the learning performance of S-GAIL with that of existing methods}
We evaluated the learning performance of S-GAIL by comparing it with existing GAIL models.
The compared models were InfoGAIL under multiple conditions: with/without the AIRL discriminator structure (described in Section \ref{sec:Situated GAIL}) and with/without the entropy-regularized coefficient (ERC) in the generator's objective function.
Namely, there were InfoGAIL, InfoGAIL+ARIL, and InfoGAIL+ARIL with the ERC.
The parameter $\beta$ for the ERC war set to 0.9, both $\alpha_{\bm w}$ and $\alpha_{\bm \phi}$ were set to 0.001, and $\gamma$ was 0.95 in this experiment.
Each model was trained five times with random initial values.

\begin{figure}[t]
\begin{center}
\includegraphics[width=0.8\columnwidth]{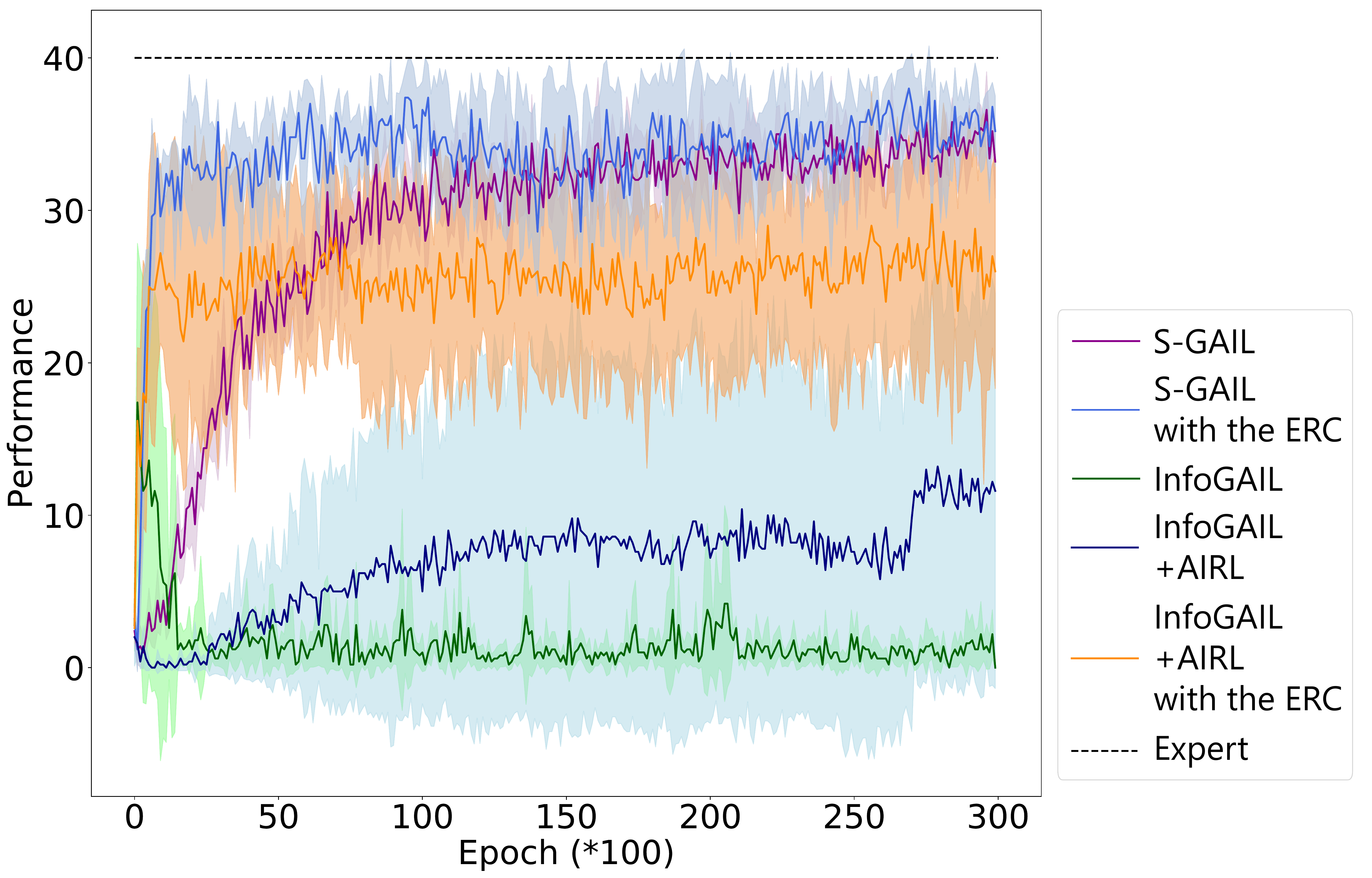}
\caption{Performance of different models. The horizontal and vertical axes respectively plot the epoch and the number of trajectories that reached the goal when 40 initial states were randomly selected and the episode was started.}
\label{fig:exp1_result1}
\end{center}
\end{figure}

\begin{figure}
    \centering
    \subfigure[S-GAIL with the ERC]{\includegraphics[width=0.3\columnwidth]{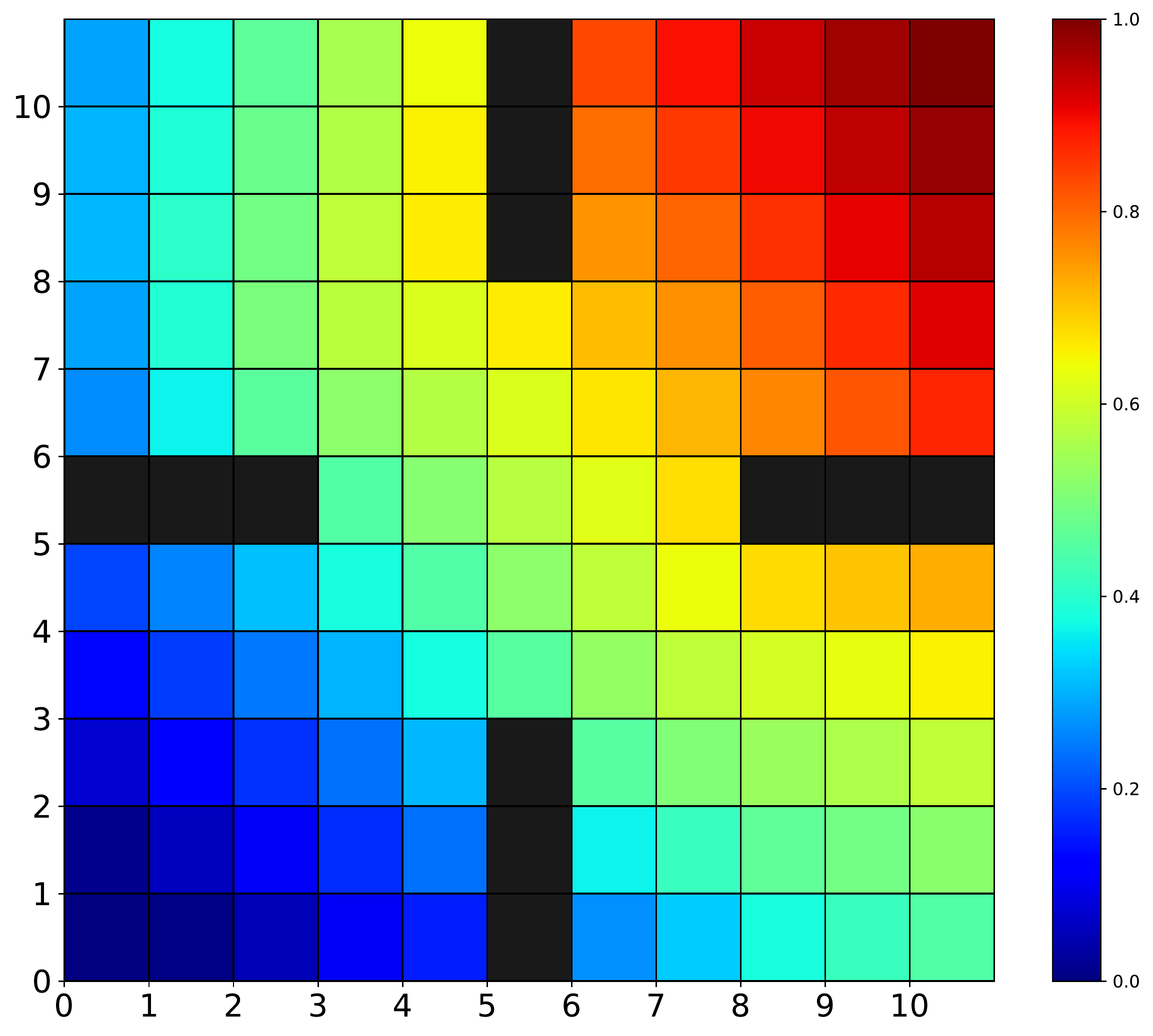}}
    \subfigure[InfoGAIL]{\includegraphics[width=0.3\columnwidth]{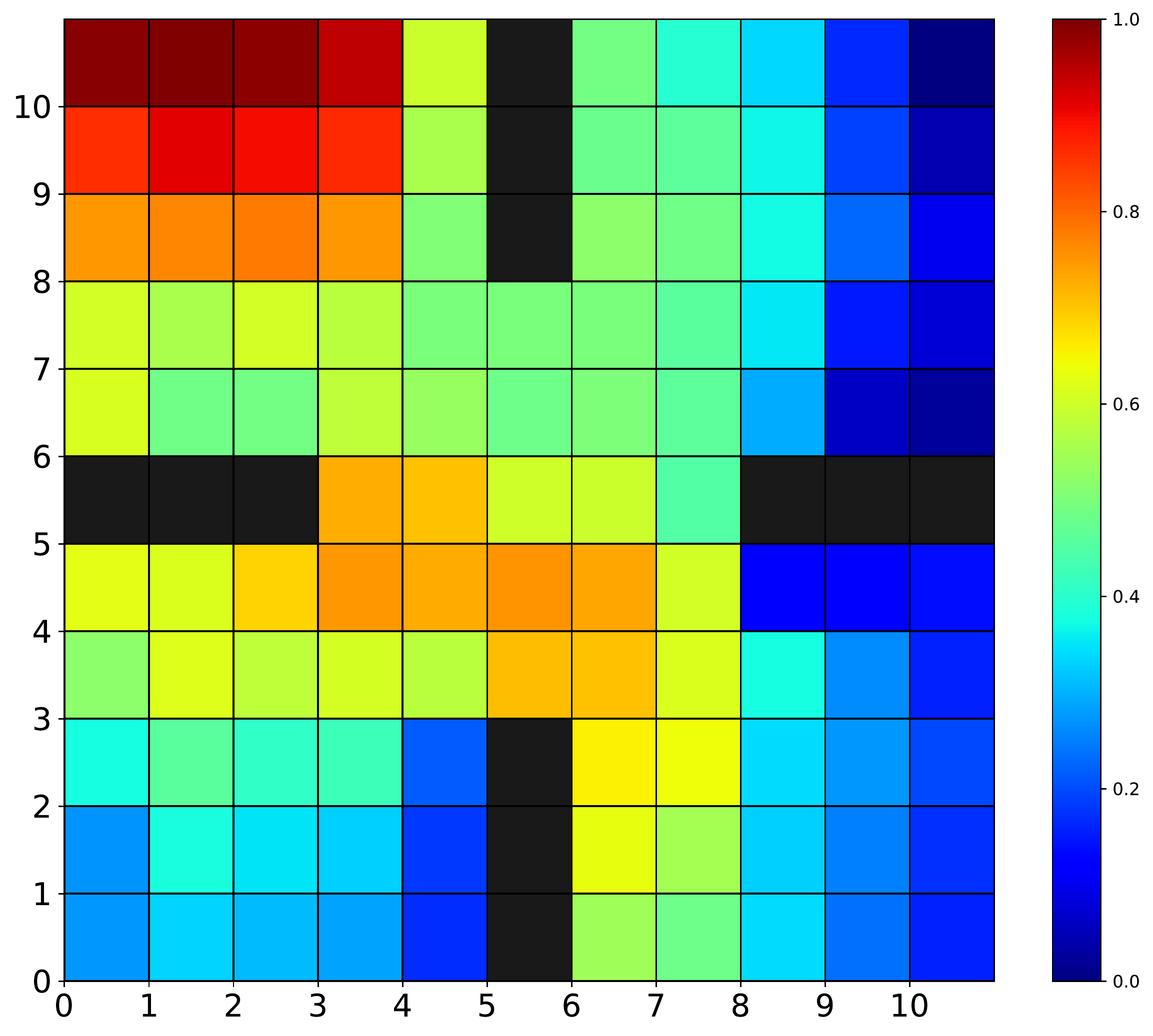}}
    \subfigure[InfoGAIL+AIRL with the ERC]{\includegraphics[width=0.3\columnwidth]{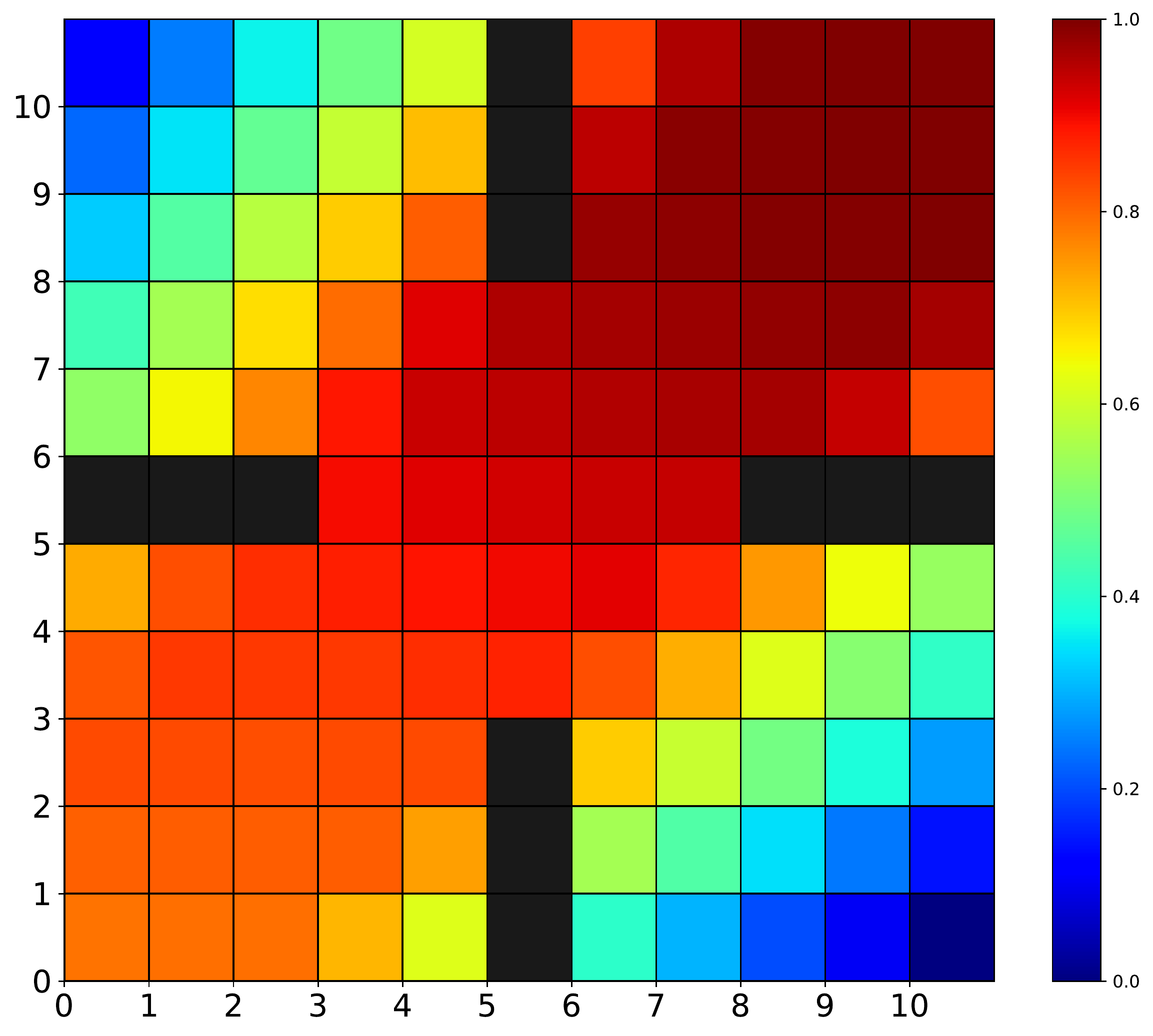}}
    \subfigure[S-GAIL with the ERC]{\includegraphics[width=0.3\columnwidth]{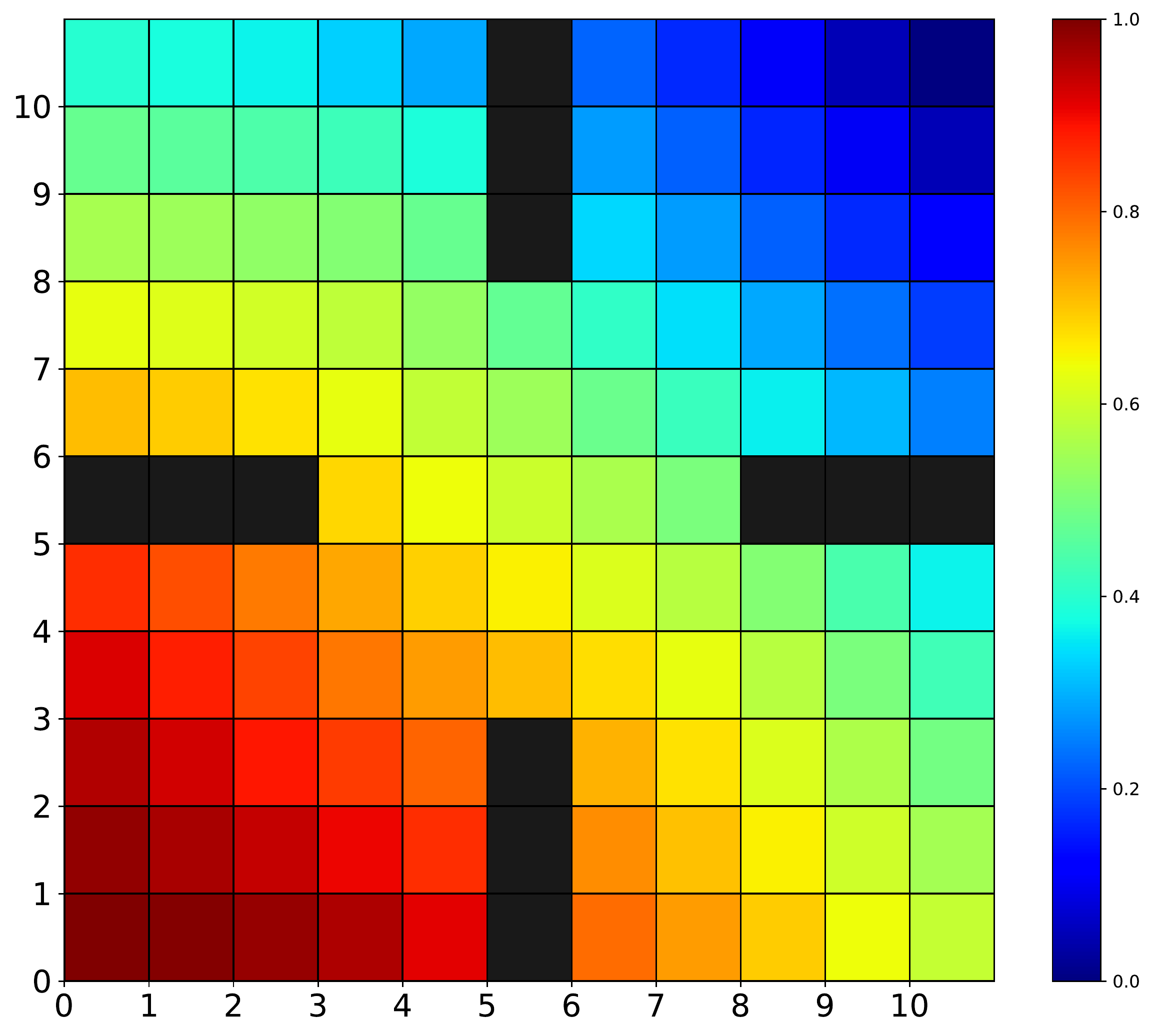}}
    \subfigure[InfoGAIL]{\includegraphics[width=0.3\columnwidth]{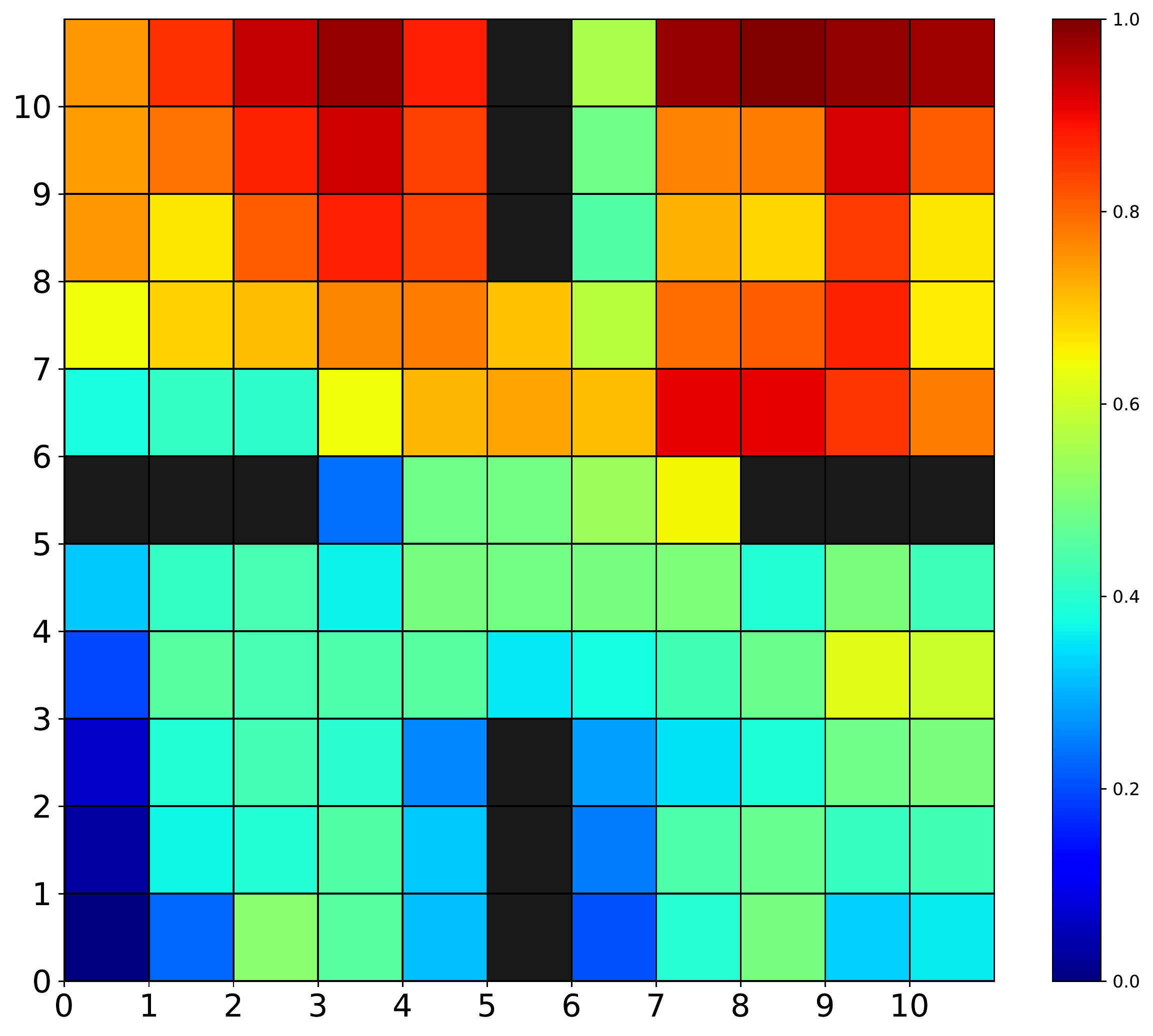}}
    \subfigure[InfoGAIL+AIRL with the ERC]{\includegraphics[width=0.3\columnwidth]{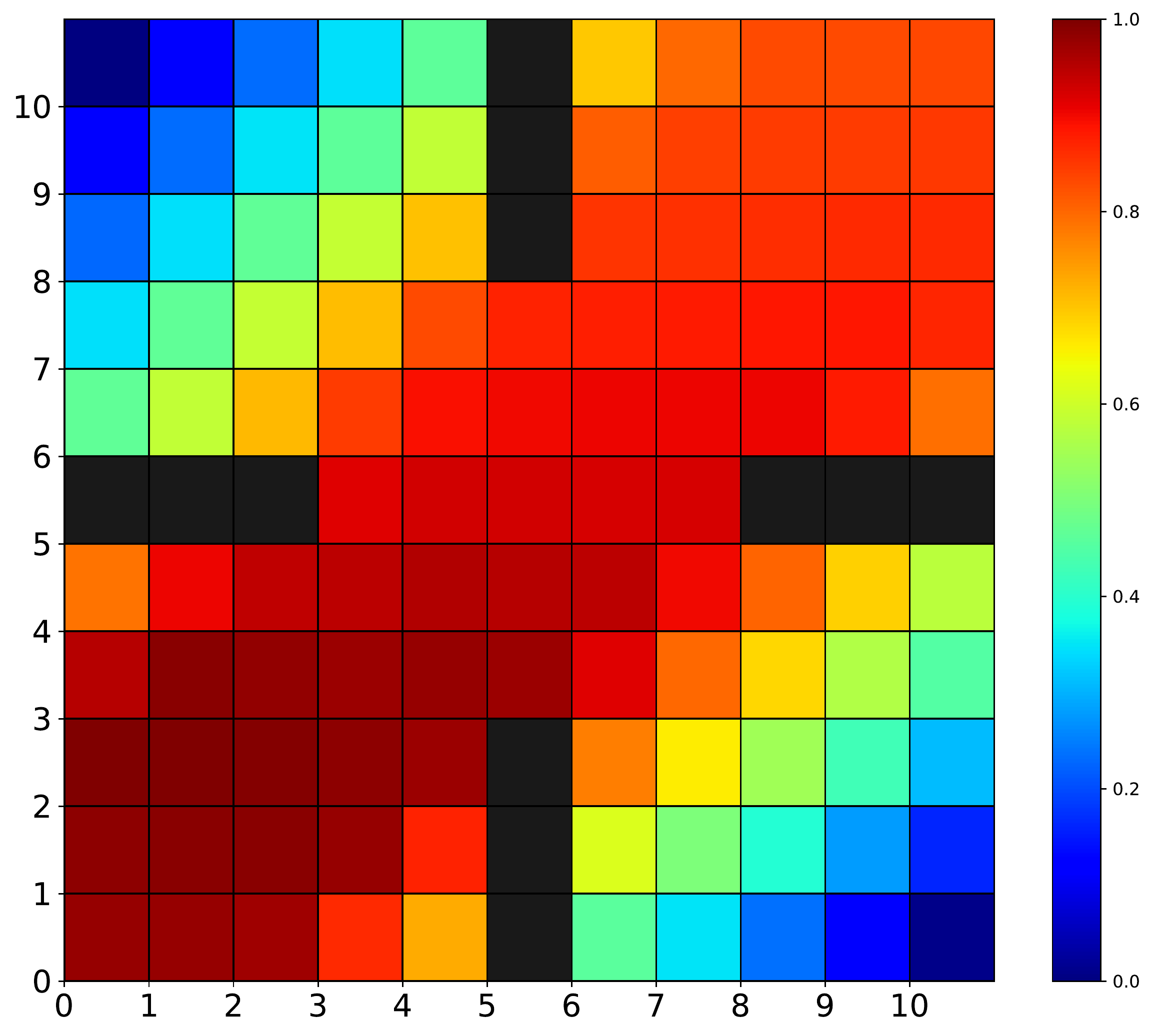}}
    \caption{State values acquired in each model. (a), (b), and (c) correspond to task 1. (d), (e), and (f) correspond to task 2.}
    \label{fig:exp1_result2}
\end{figure}

Figure \ref{fig:exp1_result1} shows the transition of the task performance over 30,000 epochs.
The performance was measured as the number of successful trials in which the agent reached the target locations among 40 trials.
Figure \ref{fig:exp1_result2} shows the value functions corresponding to task 1 (top) and task 2 (bottom).

As shown in Figure \ref{fig:exp1_result1}, S-GAIL outperformed InfoGAIL and InfoGAIL+AIRL regardless of the usage of the ERC.
By comparing the performance and value function of S-GAIL with those of InfoGAIL, the differences were clarified by considering the odds ratio structure of AIRL's discriminator (equation \ref{eq:D_sGAIL}).
The value functions of S-GAIL (Figures \ref{fig:exp1_result2}(a) and (d)) differed from those of InfoGAIL (Figures \ref{fig:exp1_result2}(b) and (e)) significantly.
In Figures \ref{fig:exp1_result2}(a) and (d), high state values were separately located at the goal positions corresponding to the tasks; however, in Figures \ref{fig:exp1_result2}(b) and (e), the high state values were confused in each value space.
The discriminator structure of AIRL with task variables (i.e., equation \ref{eq:D_sGAIL}) enabled the model to recover the two independent rewards from the expert's demonstrations.

Nevertheless, InfoGAIL+AIRL and InfoGAIL+AIRL with the ERC showed better performance than that of InfoGAIL.
In particular, InfoGAIL+AIRL with the ERC did not completely fail to learn the two tasks but achieved approximately 70\% of the expert performance.
The value functions of this model (Figures \ref{fig:exp1_result2}(c) and (f)) appeared to be separated, corresponding to each task.
The reason why InfoGAIL+AIRL and InfoGAIL+AIRL with the ERC models outperformed InfoGAIL is that they could condition the output of the discriminator (i.e., reward function) via the task variable  $c$ of the policy function $\pi_{\bm \theta}(a|s,c)$:
\begin{align}
  D_{\bm w} &= \frac{\exp(f_{\bm w}(s,a))}{\exp(f_{\bm w}(s,a))+\pi_{\bm \theta}(a|s,c)} \nonumber & \hat{R} &= f_{\bm w}(s,a) - (1 - \beta) \log \pi_{\bm \theta}(a|s,c). \nonumber
\end{align}
The above structure of the discriminator based on AIRL could represent the reward function.
Namely, the discriminator of InfoGAIL+AIRL and InfoGAIL+AIRL with the ERC models estimated different reward functions according to the policies $\pi_{\bm \theta}(a|s,c)$, which are conditioned by the task variable $c$.
However, the conditioning of reward functions by the task variable in the InfoGAIL+AIRL model was weaker than that in S-GAIL.
S-GAIL directly integrated the task variable into $f_{\bm w}(s,a,c)$ and $\log \pi_{\bm \theta}(a|s,c)$ by introducing the variable into both the discriminator and generator.
In this experiment, the target locations of the two tasks were positioned diagonally.
In other words, the two reward function conditioned by the task variable $c$ were acquired with marginal interference.
Therefore, InfoGAIL+AIRL with the ERC could incorporate the task variable $c$ (Figures \ref{fig:exp1_result2}(c) and (f)) in the pseudoreward $\hat{R}$.

Comparing the results with and without the ERC suggests that the ERC appropriately balanced $f_{\bm w}(s,a)$ and $\log \pi_{\bm \theta}(a|s,c)$ under both S-GAIL and InfoGAIL+AIRL conditions (as shown in Figure 4).
Under the S-GAIL settings, both S-GAIL and S-GAIL with the ERC models achieved similar performance at the end of training; however, the convergence was significantly faster in S-GAIL with the ERC than in S-GAIL.
The parameter $\beta$ was set to 0.9 in this experiment; that is, the entropy term of the policy was strongly omitted.
This balance of the objective function of the generator led to quicker convergence.
To evaluate the effect of the ERC in more detail, we conducted an additional experiment as described in the next section.

\subsubsection{Result 2: Effect of entropy regularization on learning}
Next, we carefully examined the effect of the ERC on learning.
The ERC helps to balance the exploration and the approximation of the expert policy and to train the generator at the early stage of training.
In this experiment, we used only five out of thirty expert trajectories for each task; thus, exploration for the generator was required to improve the performance.
The effect of the ERC was validated under four $\beta$ conditions: 1) $\beta=0.9$, 2) $\beta=0.6$, 3) $\beta$ was changed from 0.9 to 0.6, and 4) $\beta$ was changed from 0.9 to 0.0.

\begin{figure}[t]
\begin{center}
\includegraphics[width=0.8\columnwidth]{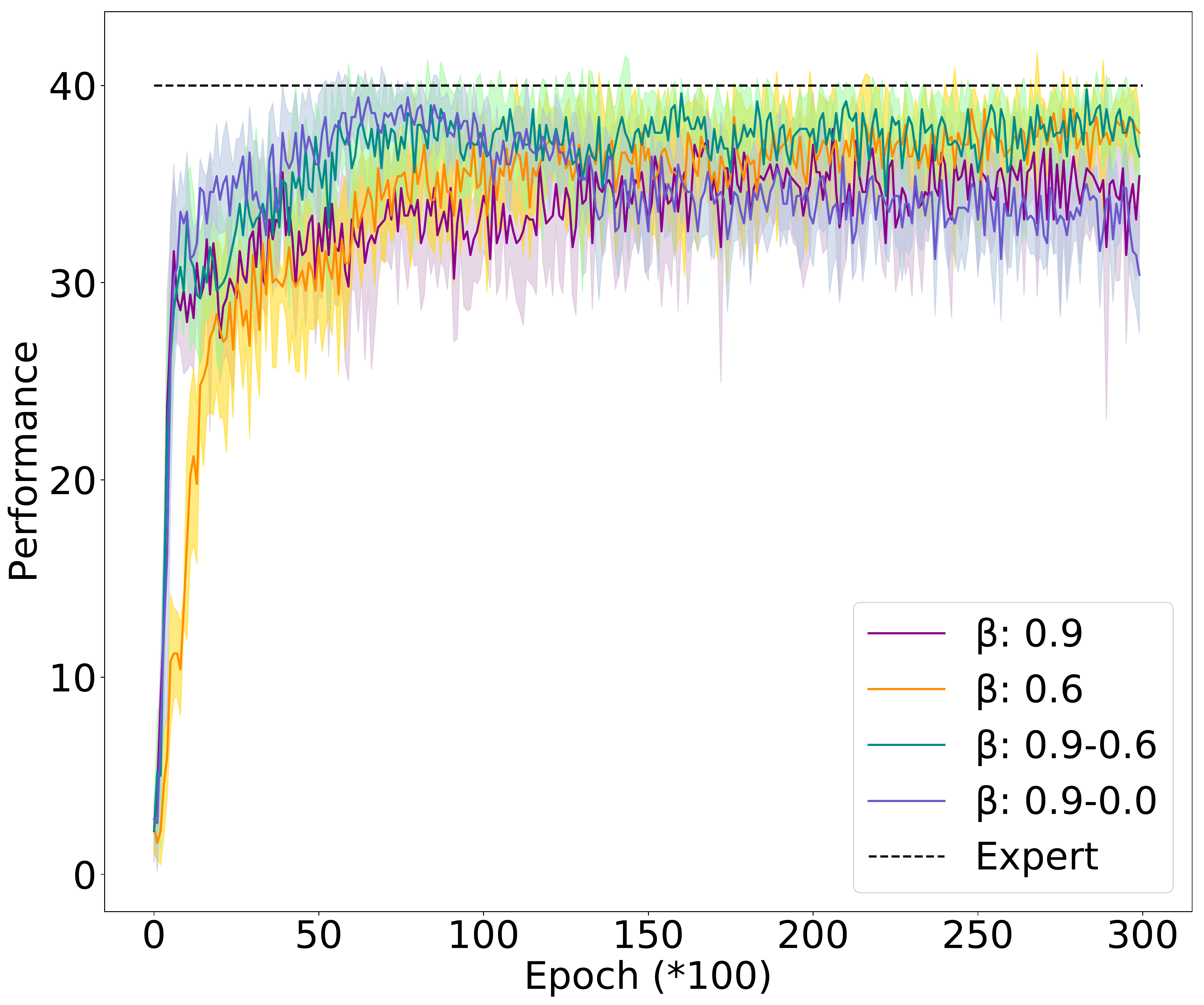}
\caption{Performance for different settings of $\beta$.}
\label{fig:exp2_result1}
\end{center}
\end{figure}

\begin{figure}
    \centering
    \subfigure[$\beta = 0.6$ at 10,000 epochs]{\includegraphics[width=0.3\columnwidth]{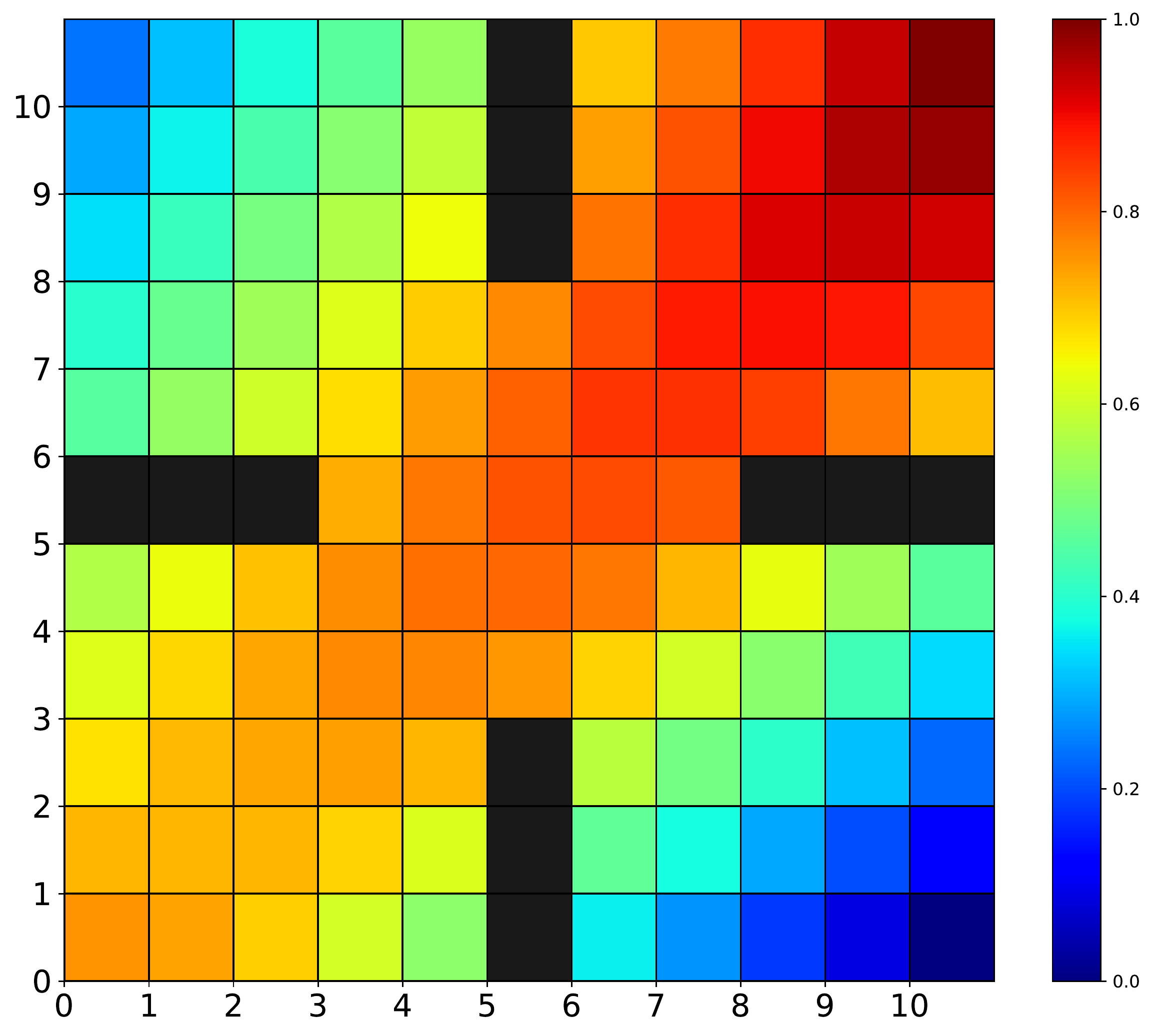}}
    \subfigure[$\beta = 0.6$ at 30,000 epochs]{\includegraphics[width=0.3\columnwidth]{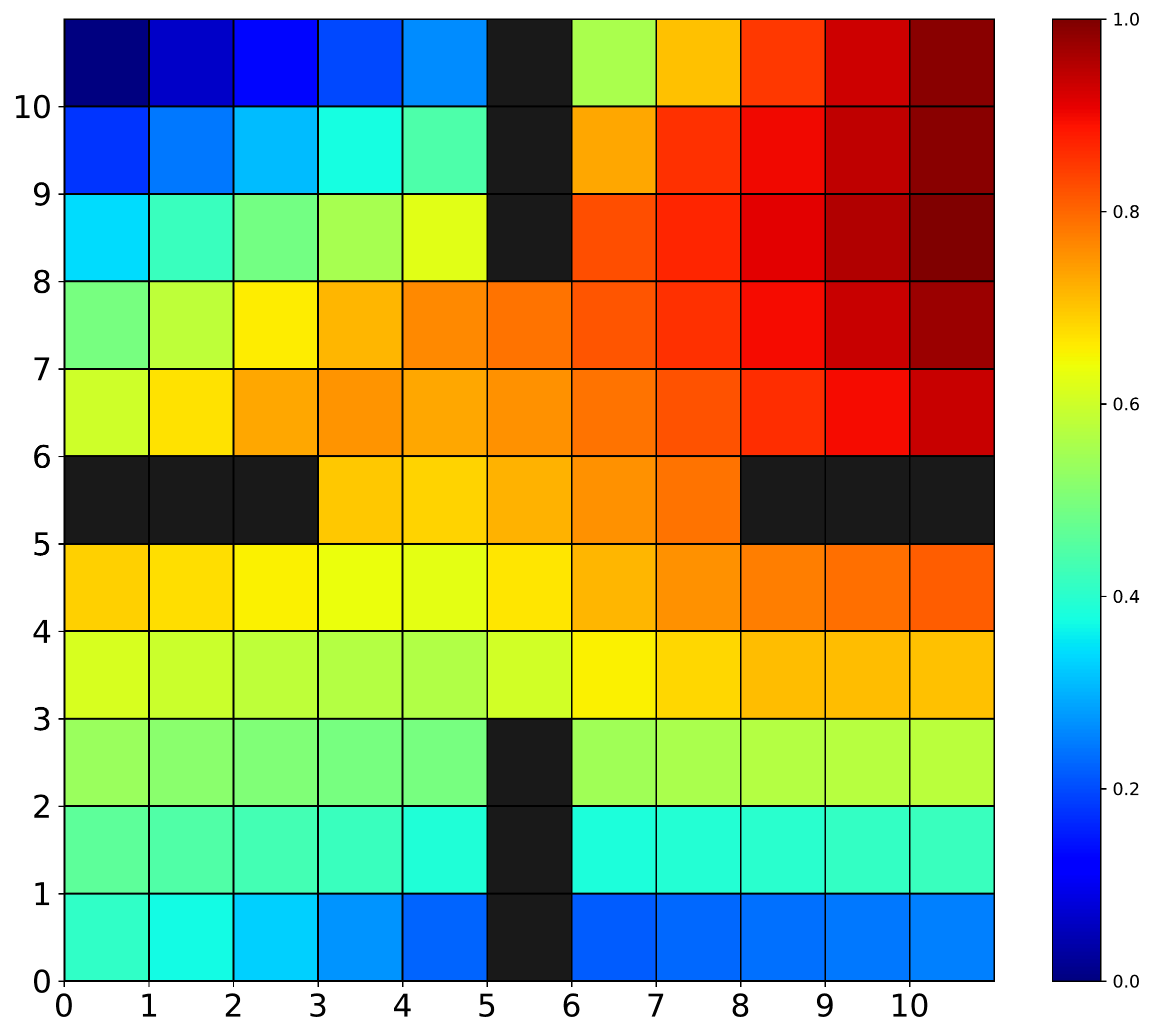}} \\
    \subfigure[$\beta = 0.9 \to 0.6$ at 10,000 epochs]{\includegraphics[width=0.3\columnwidth]{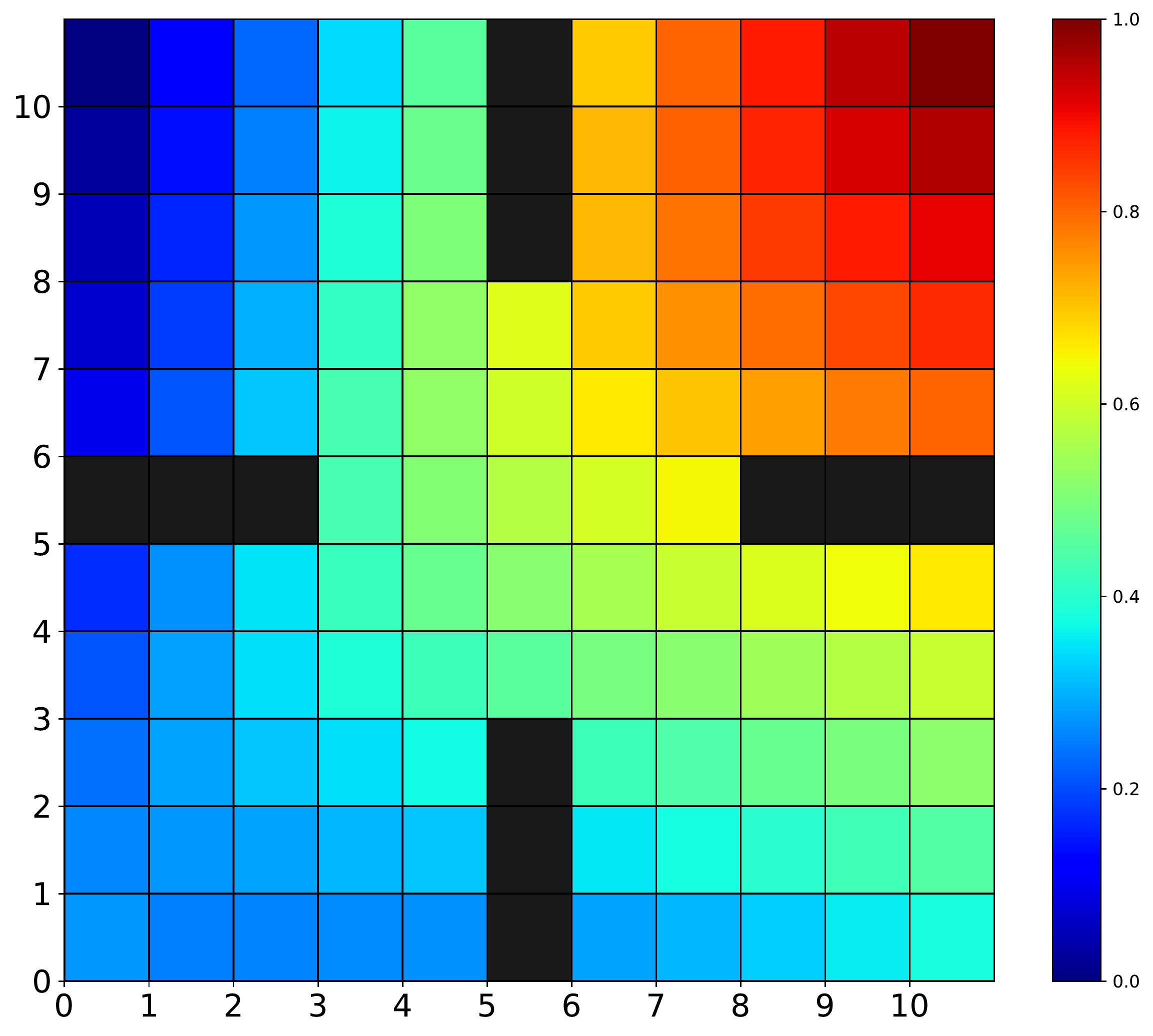}}
    \subfigure[$\beta = 0.9 \to 0.6$ at 30,000 epochs]{\includegraphics[width=0.3\columnwidth]{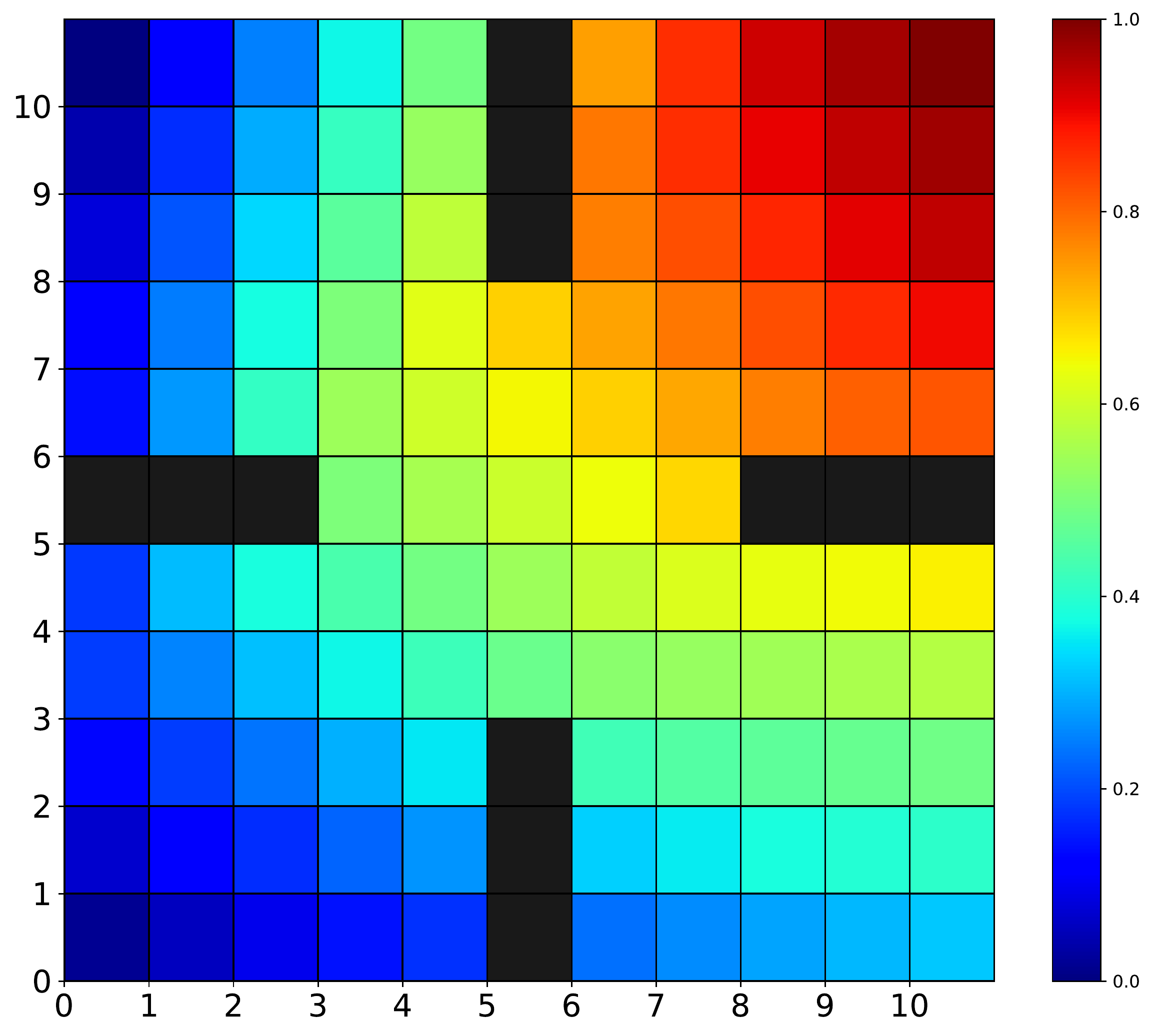}}
    \caption{Value functions under the conditions $\beta = 0.6$ and $\beta = 0.9 \to 0,6$ for task 1.}
    \label{fig:exp2_result2}
\end{figure}

Figure \ref{fig:exp2_result1} shows the influence of the ERC under the four conditions.
The comparison of the first two conditions (i.e., $\beta$ = 0.9 and 0.6) demonstrated that a smaller value of $\beta$ ($\beta$ = 0.6) resulted in slower but greater improvement in performance owing to more explorations.
Adjusting $\beta$ through learning enabled the model to take advantage of the above two fixed conditions.
The model with $\beta = 0.9 \to 0.6$ achieved both higher performance and earlier convergence.
Figure \ref{fig:exp2_result2} shows the value functions under the conditions of $\beta=0.6$ and $\beta=0.9 \to 0.6$ at 10,000 and 30,000 learning epochs.
At 10,000 learning epochs, high state values were widely distributed in the environment for $\beta=0.6$ (Figure \ref{fig:exp2_result2}(a)), whereas high values were already associated with the goal position under the $\beta=0.9 \to 0.6$ condition (Figure \ref{fig:exp2_result2}(c)).
Only in the later phase of learning, the high state values moved close to the goal position for $\beta=0.6$ (figure \ref{fig:exp2_result2}(b)) owing to the exploration.
In the case of $\beta=0.9 \to 0.0$, the performance was quickly improved in the early stage; however, it decreased as learning progressed.
This reduction was caused by a higher entropy of the policy.
Because the high entropy led to a uniform distribution of the policy function, the action selection of the agent may become noisy.
Thus, we suggest that the proper setting of $\beta$ is important to benefit from this mechanism.

\subsubsection{Result 3: Comparison with a single-task learning model}
Finally, we evaluated the advantage of using a single model for learning multiple tasks.
A potential advantage is to share acquired knowledge among tasks (i.e., policies and dynamics of the environment).
It is thus expected to reduce the number of demonstrations and the amount of training because of the common knowledge.

In this experiment, we designed three conditions for learning the two tasks:
\begin{enumerate}
  \item single S-GAIL with a scheduled ERC (i.e., $\beta=0.9 \to 0.6$) (S-GAIL with the ERC: double task).
  \item two AIRL models learning the two tasks separately (AIRL: single task).
  \item two AIRL models learning the two tasks separately using a scheduled ERC  (i.e., $\beta=0.9 \to 0.6$) (AIRL with the ERC: single task).
\end{enumerate}
Again, we used only five expert trajectories for each task.
Under all conditions, each model learned the same number of demonstrations by the expert.
That is, S-GAIL learned a half of the number of demonstrations for each task from other conditions.

\begin{figure}[t]
\begin{center}
\includegraphics[width=0.8\columnwidth]{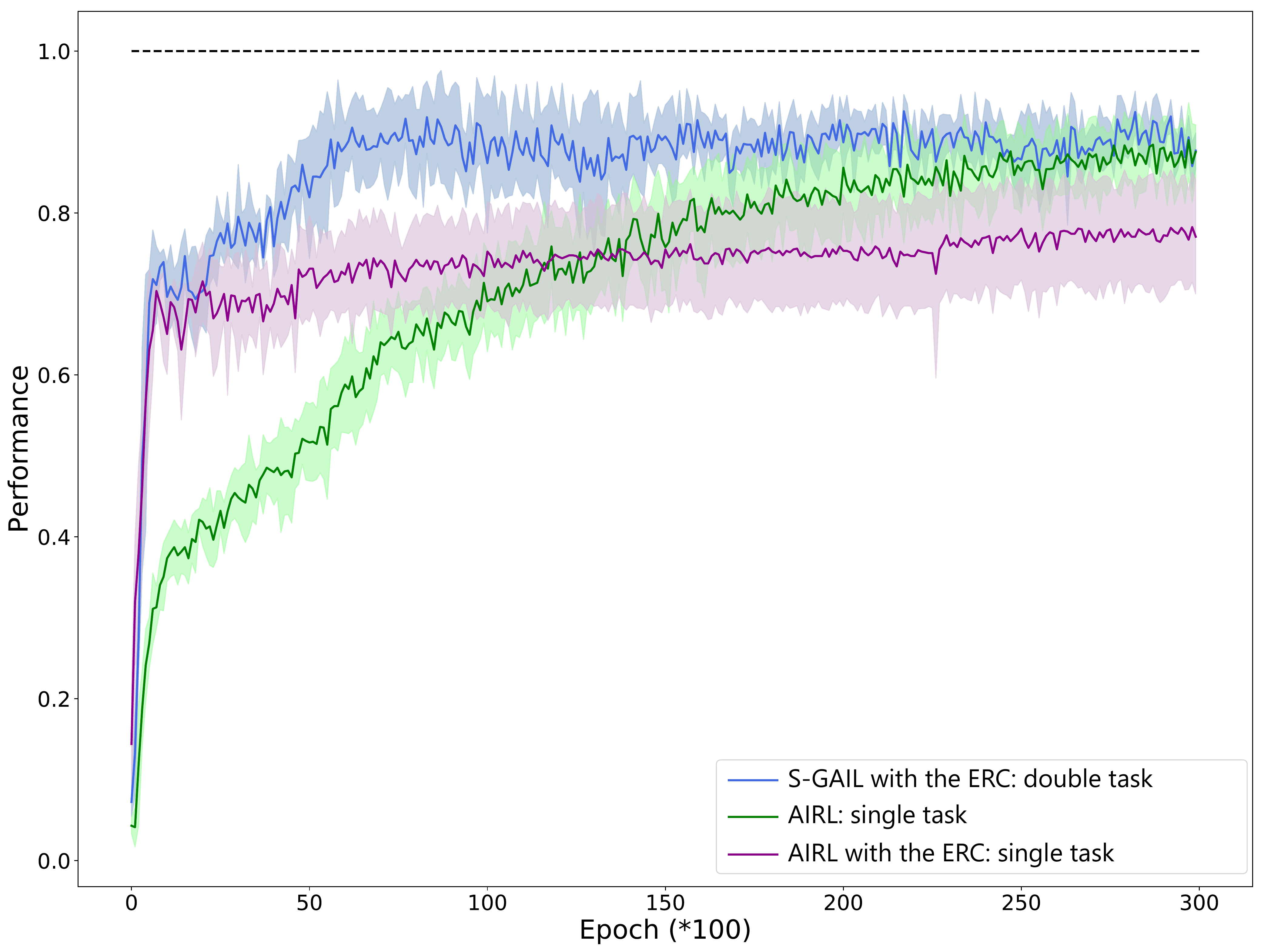}
\caption{Performance of different models and settings.}
\label{fig:exp3}
\end{center}
\end{figure}

Figure \ref{fig:exp3} shows the performance of task 1 under the three conditions.
We allow the agent to start from all possible states except the goal state.
S-GAIL outperformed the other models, although the number of training was half of the other conditions.
Furthermore, S-GAIL acquired the policy faster than the others owing to the ERC.
The reason why S-GAIL achieved better performance is that it shared the model parameters for different policies.
In particular, it benefited from sharing the generator's parameters because the state and action spaces and the transition probability were common among the two tasks.
To summarize, S-GAIL is advantageous in that it can acquire multiple policies more quickly by introducing task variables for both the discriminator and generator and the ERC for the objective function.

\subsection{Robot-arm simulator}
\begin{figure}[t]
\begin{center}
  \includegraphics[width=0.45\columnwidth]{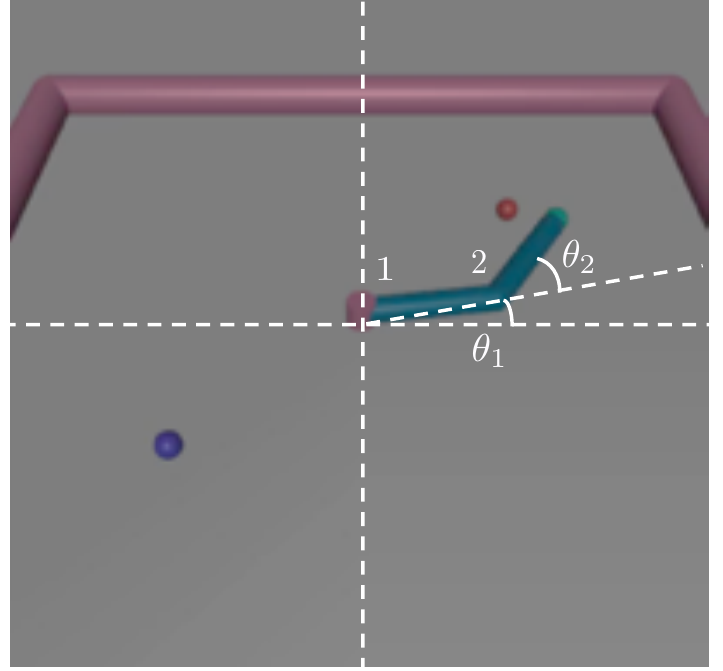}
  \caption{Environment of the Reacher robot.}
\label{fig:reacher}
\end{center}
\end{figure}

The second experiment was conducted using a robot-arm simulator.
In this experiment, we examined whether the proposed method can learn to imitate robot-arm reaching behavior in a continuous space.
We used the Reacher-v2 environment provided by the OpenAI Gym platform using the MuJoCo physical simulator \cite{Brockman_Openai2016}.
Figure \ref{fig:reacher} shows the environment of Reacher.
The robot manipulator has two movable joints.
Link 1 was fixed at the coordinates $(0,0)$, and Link 2 was attached to Link 1.
The robot aimed to reach for two target objects placed at the top-right corner (red ball) and bottom-left corner (blue ball) in the environment.
The state of the robot $s$ was defined by a continuous six-dimensional vector $\sin \theta_i$, $\cos \theta_i$, $\dot{\theta_i}$, where $i$ is the link number ($i=1,2$) and $\theta_i$ is the relative angle of the link in the $x$-$y$ plane.
The initial states of the robot were sampled as $\theta_1 \sim \mathcal{U}(-3.0, 1.3) {\rm rad}$, $\theta_2 = 0.0 {\rm rad}$  where $\mathcal{U}$ indicates the uniform distribution.
We designed the expert's behaviors so that it moves the tip from the initial state to one of the target objects while taking the shortest path.

\begin{figure}
    \centering
    \subfigure[Performance]{\includegraphics[width=0.4\columnwidth]{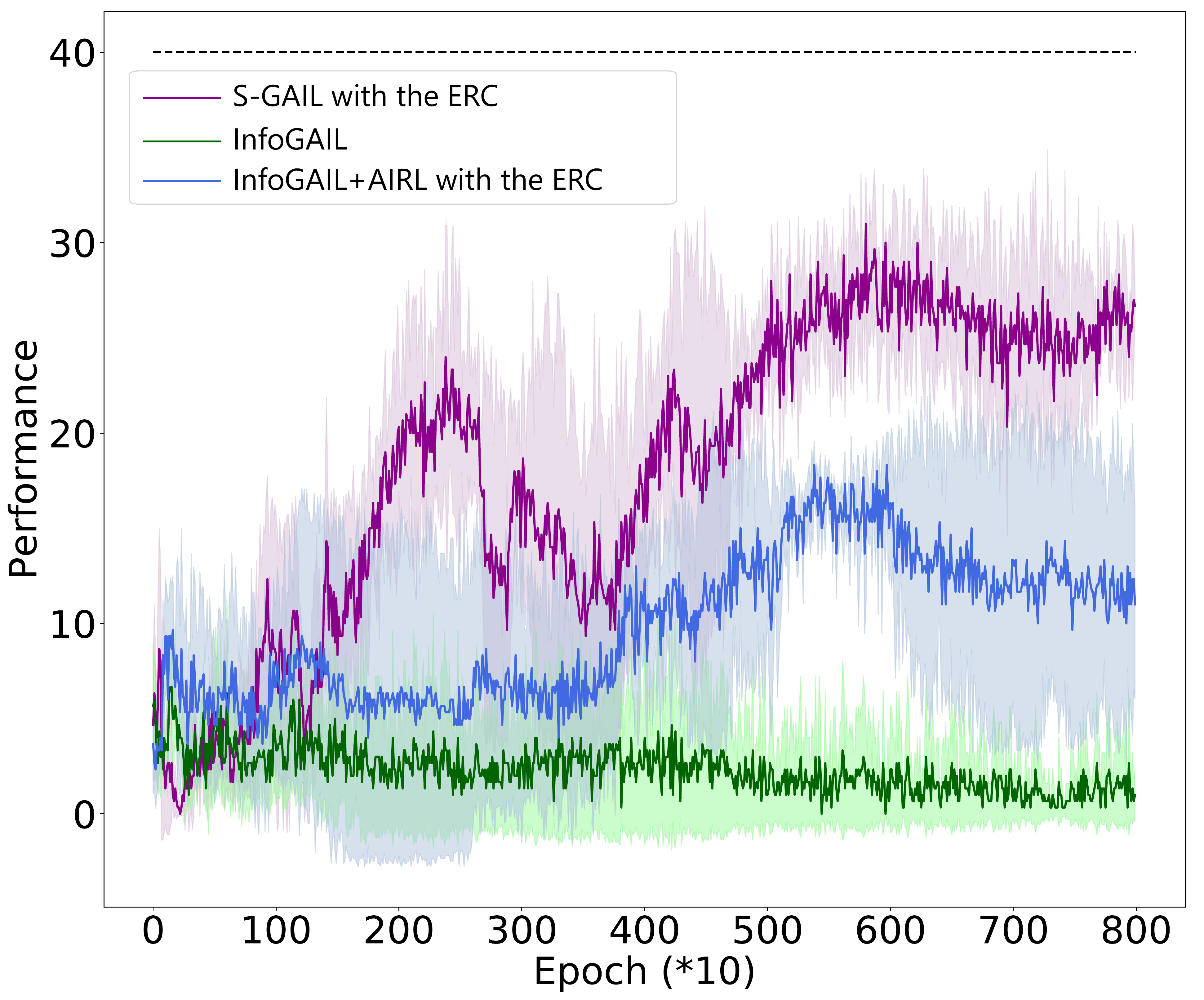}}
    \subfigure[Breakdown of tasks performed using S-GAIL]{\includegraphics[width=0.4\columnwidth]{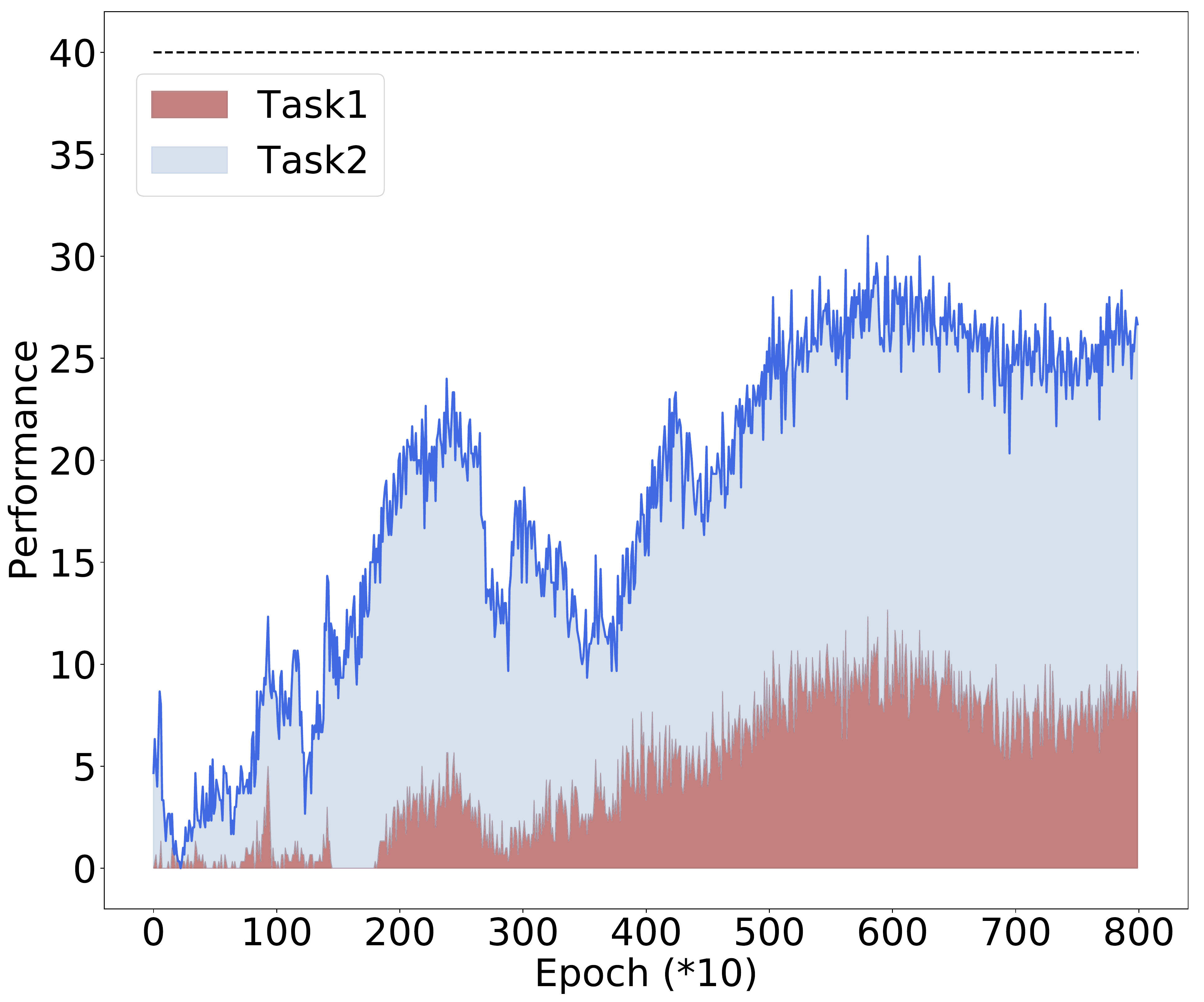}}
    \caption{Performance of different models. The horizontal and vertical axes respectively plot the epoch and the number of trajectories that reached the goal when the initial state was randomly selected and the episode started. (a)Performance and (b) breakdown of the tasks performed using S-GAIL.}
    \label{fig:result_reacher}
\end{figure}

Figures \ref{fig:result_reacher}(a) and (b) show the learning curves.
They plot the number of successful trials among 40 trials with different initial conditions.
We again compared our method to InfoGAIL with/without AIRL.
Figure \ref{fig:result_reacher}(a) shows that S-GAIL exhibits the highest performance.
Figure \ref{fig:result_reacher}(b) shows the breakdown of the tasks performed by the proposed method; the results indicate that S-GAIL simultaneously learned the two tasks.

\begin{figure}
    \centering
    \subfigure[Task 1: Reaching for the red ball]{\includegraphics[width=0.4\columnwidth]{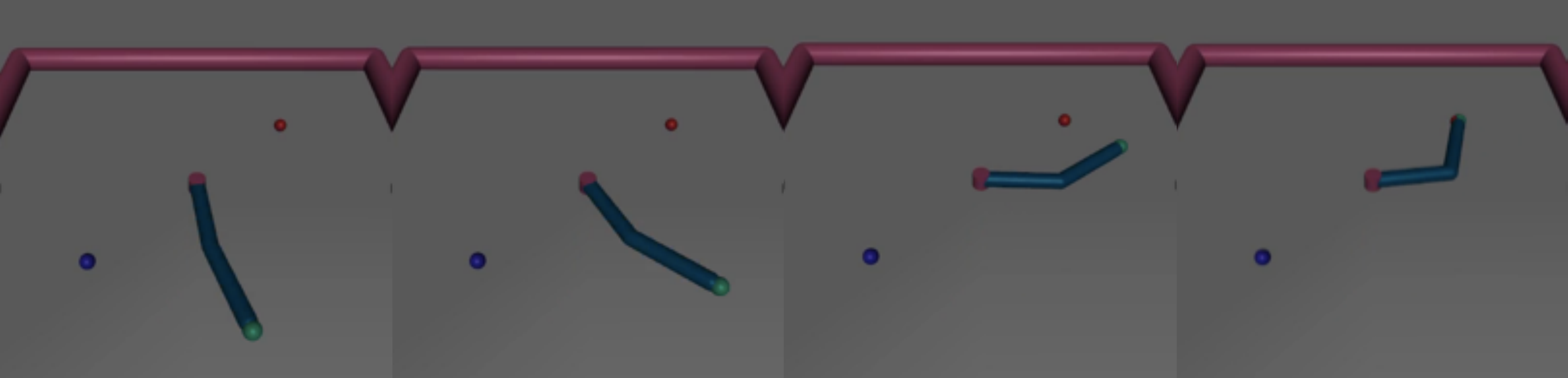}}
    \subfigure[Task 2: Reaching for the blue ball]{\includegraphics[width=0.4\columnwidth]{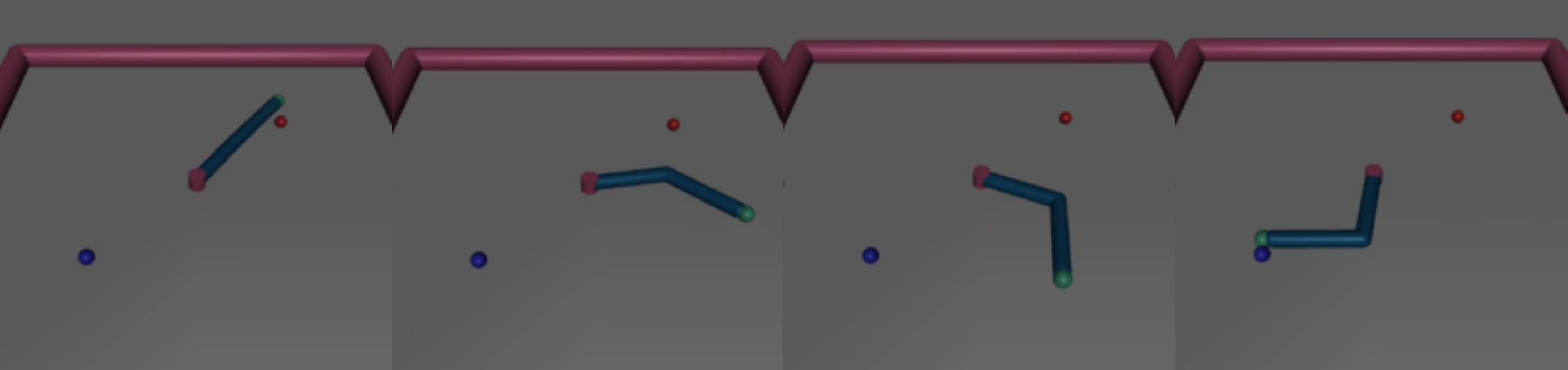}}
    \caption{Trajectory of the robot arm for each task.}
    \label{fig:result_reacher2}
\end{figure}

Figure \ref{fig:result_reacher2} presents snapshots of the robot while reaching for (a) the red ball and (b) the blue ball.
The proposed method enabled the manipulator to smoothly reach for the object corresponding to each task.
In particular, the robot in Figure \ref{fig:result_reacher2}(b) began at an initial state near the red object; however, it successfully moved its tip to the blue object without being confused by the red one.
This example demonstrated that S-GAIL could independently estimate a reward function for each task.

\section{Conclusion}
\label{sec:conclusion}
We proposed S-GAIL, which extends GAIL to take advantage of both InfoGAIL and AIRL.
The task variables and a specific form of the discriminator enabled the framework to estimate multiple reward functions and policies corresponding to multiple tasks.
Owing to the shared representations among multiple tasks, S-GAIL achieved faster and better learning compared to existing frameworks. 
It also guaranteed convergence to the optimal solution as in GAIL models.
Furthermore, we introduced a coefficient-correction entropy regularization term to the objective function of the generator.
This achieved a trade-off between the speed of learning convergence and performance by switching from the maximization of the estimated reward to the maximization of the entropy in policy learning.

In the future, we plan to validate S-GAIL under a real robot condition where the robot learns actions from human teleoperation.
In such a case, S-GAIL can use task switching signals and language instruction of task names (e.g., "now, cleaning the room", "grasping ball", etc.) from operators as task variables $c$.
We will also attempt to introduce continuous variables with current discrete task variables to modulate the shape of reward functions.
We also intend to extend the current method to automatically estimate the number and type of tasks contained in experts' behaviors.


\section*{Acknowledgement}
We would like to thank Dr. Eiji Uchibe for useful discussions.
This work was supported by JST CREST project 'Cognitive Mirroring' (Grant Number: JPMJCR16E2), Japan.


\bibliographystyle{tADR}
\bibliography{example}  
\end{document}